# AntNet: Distributed Stigmergetic Control for Communications Networks

**Gianni Di Caro**                                         GDICARO@IRIDIA.ULB.AC.BE
**Marco Dorigo**                                              MDORIGO@ULB.AC.BE
*IRIDIA, Université Libre de Bruxelles*
*50, av. F. Roosevelt, CP 194/6, 1050 - Brussels, Belgium*

## Abstract

This paper introduces AntNet, a novel approach to the adaptive learning of routing tables in communications networks. AntNet is a distributed, mobile agents based Monte Carlo system that was inspired by recent work on the ant colony metaphor for solving optimization problems. AntNet's agents concurrently explore the network and exchange collected information. The communication among the agents is indirect and asynchronous, mediated by the network itself. This form of communication is typical of social insects and is called *stigmergy*. We compare our algorithm with six state-of-the-art routing algorithms coming from the telecommunications and machine learning fields. The algorithms' performance is evaluated over a set of realistic testbeds. We run many experiments over real and artificial IP datagram networks with increasing number of nodes and under several paradigmatic spatial and temporal traffic distributions. Results are very encouraging. AntNet showed superior performance under all the experimental conditions with respect to its competitors. We analyze the main characteristics of the algorithm and try to explain the reasons for its superiority.

## 1. Introduction

Worldwide demand and supply of communications networks services are growing exponentially. Techniques for network control (i.e., online and off-line monitoring and management of the network resources) play a fundamental role in best exploiting the new transmission and switching technologies to meet user's requests.

Routing is at the core of the whole network control system. Routing, in conjunction with the admission, flow, and congestion control components, determines the overall network performance in terms of both quality and quantity of delivered service (Walrand & Varaiya, 1996). Routing refers to the distributed activity of building and using *routing tables*, one for each node in the network, which tell incoming data packets which outgoing link to use to continue their travel towards the destination node.

Routing protocols and policies have to accommodate conflicting objectives and constraints imposed by technologies and user requirements rapidly evolving under commercial and scientific pressures. Novel routing approaches are required to efficiently manage distributed multimedia services, mobile users and networks, heterogeneous inter-networking, service guarantees, point-to-multipoint communications, etc. (Sandick & Crawley, 1997; The ATM Forum, 1996).

The adaptive and distributed routing algorithm we propose in this paper is a mobile-agent-based, online Monte Carlo technique inspired by previous work on artificial ant colonies and, more generally, by the notion of *stigmergy* (Grassé, 1959), that is, the indirect communication taking place among individuals through modifications induced in their environment.

Algorithms that take inspiration from real ants' behavior in finding shortest paths (Goss, Aron, Deneubourg, & Pasteels, 1989; Beckers, Deneubourg, & Goss, 1992) using as information only the trail of a chemical substance (called *pheromone*) deposited by other ants, have recently been successfully applied to several discrete optimization problems (Dorigo, Maniezzo, & Colorni, 1991; Dorigo, 1992; Dorigo, Maniezzo, & Colorni, 1996; Dorigo & Gambardella, 1997; Schoonderwoerd, Holland, Bruten, & Rothkrantz, 1996; Schoonderwoerd, Holland, & Bruten, 1997; Costa & Hertz, 1997). In all these algorithms a set of artificial ants collectively solve the problem under consideration through a cooperative effort. This effort is mediated by indirect communication of information on the problem structure the ants concurrently collect while building solutions by using a stochastic policy. Similarly, in AntNet, the algorithm we propose in this paper, a set of concurrent distributed agents collectively solve the adaptive routing problem. Agents adaptively build routing tables and local models of the network status by using indirect and non-coordinated communication of information they collect while exploring the network.

To ensure a meaningful validation of our algorithm performance we devised a realistic simulation environment in terms of network characteristics, communications protocol and traffic patterns. We focus on IP (Internet Protocol) datagram networks with irregular topology and consider three real and artificial topologies with an increasing number of nodes and several paradigmatic temporal and spatial traffic distributions. We report on the behavior of AntNet as compared to some effective static and adaptive state-of-the-art routing algorithms (vector-distance and link-state shortest paths algorithms (Steenstrup, 1995), and recently introduced algorithms based on machine learning techniques).

AntNet shows the best performance and the most stable behavior for all the considered situations. In many experiments its superiority is striking. We discuss the results and the main properties of our algorithm, as compared with its competitors.

The paper is organized as follows. In Section 2 the definition, taxonomy and characteristics of the routing problem are reported. In Section 3 we describe the communication network model we used. Section 4 describes in detail AntNet, our novel routing algorithm, while in Section 5 we briefly describe the algorithms with which we compared AntNet. In Section 6, the experimental settings are reported in terms of traffic, networks and algorithm parameters. Section 7 reports several experimental results. In Section 8 we discuss these results and try to explain AntNet's superior performance. Finally, in Section 9, we discuss related work, and in Section 10, we draw some conclusions and outline directions for future research.

## 2. Routing: Definition and Characteristics

Routing in distributed systems can be characterized as follows. Let $G = (V, E)$ be a directed weighted graph, where each node in the set $V$ represents a processing/queuing and/or forwarding unit and each edge is a transmission system. The main task of a routing algorithm is to direct data flow from source to destination nodes maximizing network performance.

In the problems we are interested in, the data flow is not statically assigned and it follows a stochastic profile that is very hard to model.

In the specific case of communications networks (Steenstrup, 1995; Bertsekas & Gallager, 1992), the routing algorithm has to manage a set of basic functionalities and it tightly interacts with the congestion and admission control algorithms, with the links' queuing policy, and with the user-generated traffic. The core of the routing functions is (i) the acquisition, organization and distribution of information about user-generated traffic and network states, (ii) the use of this information to generate feasible routes maximizing the performance objectives, and (iii) the forwarding of user traffic along the selected routes.

The way the above three functionalities are implemented strongly depends on the underlying network switching and transmission technology, and on the features of the other interacting software layers. Concerning point (iii), two main forwarding paradigms are in use: *circuit* and *packet-switching* (also indicated with the terms *connection-oriented* and *connection-less*). In the circuit-switching approach, a setup phase looks for and reserves the resources that will be assigned to each incoming session. In this case, all the data packets belonging to the same session will follow the same path. Routers are required to keep state information about active sessions. In the packet-switching approach, there is no reservation phase, no state information is maintained at routers and data packets can follow different paths. In each intermediate node an autonomous decision is taken concerning the node's outgoing link that has to be used to forward the data packet toward its destination.

In the work described in this paper, we focus on the packet-switching paradigm, but the technique developed here can be used also to manage circuit-switching and we expect to have qualitatively similar results.

## 2.1 A Broad Taxonomy

A common feature of all the routing algorithms is the presence in every network node of a data structure, called *routing table*, holding all the information used by the algorithm to make the local forwarding decisions. The routing table is both a local database and a local model of the global network status. The type of information it contains and the way this information is used and updated strongly depends on the algorithm's characteristics. A broad classification of routing algorithms is the following:

- centralized *versus* distributed;
- static *versus* adaptive.

In *centralized* algorithms, a main controller is responsible for updating all the node's routing tables and/or to make every routing decision. Centralized algorithms can be used only in particular cases and for small networks. In general, the delays necessary to gather information about the network status and to broadcast the decisions/updates make them infeasible in practice. Moreover, centralized systems are not fault-tolerant. In this work, we will consider exclusively distributed routing.

In *distributed* routing systems, the computation of routes is shared among the network nodes, which exchange the necessary information. The distributed paradigm is currently used in the majority of network systems.

In *static* (or *oblivious*) routing systems, the path taken by a packet is determined only on the basis of its source and destination, without regard to the current network state. This path is usually chosen as the shortest one according to some cost criterion, and it can be changed only to account for faulty links or nodes.

*Adaptive* routers are, in principle, more attractive, because they can adapt the routing policy to time and spatially varying traffic conditions. As a drawback, they can cause oscillations in selected paths. This fact can cause circular paths, as well as large fluctuations in measured performance. In addition, adaptive routing can lead more easily to inconsistent situations, associated with node or link failures or local topological changes. These stability and inconsistency problems are more evident for connection-less than for connection-oriented networks (Bertsekas & Gallager, 1992).

Another interesting way of looking at routing algorithms is from an optimization perspective. In this case the main paradigms are:

- minimal routing *versus* non-minimal routing;
- optimal routing *versus* shortest path routing.

*Minimal* routers allow packets to choose only minimal cost paths, while *non-minimal* algorithms allow choices among all the available paths following some heuristic strategies (Bolding, Fulgham, & Snyder, 1994).

*Optimal routing* has a network-wide perspective and its objective is to optimize a function of all individual link flows (usually this function is a sum of link costs assigned on the basis of average packet delays) (Bertsekas & Gallager, 1992).

*Shortest path routing* has a source-destination pair perspective: there is no global cost function to optimize. Its objective is to determine the shortest path (minimum cost) between two nodes, where the link costs are computed (statically or adaptively) following some statistical description of the link states. This strategy is based on individual rather than group rationality (Wang & Crowcroft, 1992). Considering the different content stored in each routing table, shortest path algorithms can be further subdivided into two classes called *distance-vector* and *link-state* (Steenstrup, 1995).

Optimal routing is static (it can be seen as the solution of a multicommodity flow problem) and requires the knowledge of all the traffic characteristics. Shortest paths algorithms are more flexible, they don't require a priori knowledge about the traffic patterns and they are the most widely used routing algorithms.

In appendix A, a more detailed description of the properties of optimal and shortest path routing algorithms is reported.

In Section 4, we introduce a novel distributed adaptive method, *AntNet*, that shares the same optimization perspective as (minimal or non-minimal) shortest path algorithms but not their usual implementation paradigms (as depicted in appendix A).

## 2.2 Main Characteristics of the Routing Problem

The main characteristics of the routing problem in communications networks can be summarized in the following way:

- *Intrinsically distributed* with strong *real-time* constraints: in fact, the database and the decision system are completely distributed over all the network nodes, and failures and status information propagation delays are not negligible with respect to the user's traffic patterns. It is impossible to get complete and up-to-date knowledge of the distributed state, that remains hidden. At each decision node, the routing algorithm can only make use of local, up-to-date information, and of non-local, delayed information coming from the other nodes.

- *Stochastic and time-varying*: the session arrival and data generation process is, in the general case, non-stationary and stochastic. Moreover, this stochastic process interacts recursively with the routing decisions making it infeasible to build a working model of the whole system (to be used for example in a dynamic programming framework).

- *Multi-objective*: several conflicting *performance measures* are usually taken into account. The most common are *throughput* (bit/sec) and *average packet delay* (sec). The former measures the quantity of service that the network has been able to offer in a certain amount of time (amount of correctly delivered bits per time unit), while the latter defines the quality of service produced at the same time. Citing Bertsekas and Gallager (1992), page 367: *"the effect of good routing is to increase throughput for the same value of average delay per packet under high offered load conditions and to decrease average delay per packet under low and moderate offered load conditions"*. Other performance measures consider the impact of the routing algorithm on the network resources in terms of memory, bandwidth and computation, and the algorithm simplicity, flexibility, etc.

- *Multi-constraint*: constraints are imposed by the underlying network technology, the network services provided and the user services requested. In general, users ask for low-cost, high-quality, reliable, distributed multimedia services available across heterogeneous static and mobile networks. Evaluating technological and commercial factors, network builders and service providers try to accommodate these requests while maximizing some profit criteria. Moreover, a high level of *fault-tolerance* and *reliability* is requested in modern high-speed networks, where user sessions can formulate precise requests for network resources. In this case, once the session has been accepted, the system should be able to guarantee that the session gets the resources it needs, under any recoverable fault event.

It is interesting to note that the above characteristics make the problem of routing belong to the class of *reinforcement learning* problems with hidden state (Bertsekas & Tsitsiklis, 1996; Kaelbling, Littman, & Moore, 1996; McCallum, 1995). A distributed system of agents, the components of the routing algorithm in each node, determine a continual and online learning of the best routing table values with respect to network's performance criteria. An exact measure of evaluation that scores forwarding decisions is not available, neither online nor in the form of a training set. Moreover, because of the distributed nature of the problem and of its constraints, the complete state of the network is hidden to each agent.

## 3. The Communication Network Model

In this paper, we focus on irregular topology connection-less networks with an IP-like network layer (in the ISO-OSI terminology) and a very simple transport layer. In particular, we focus on wide-area networks (WAN). In these cases, *hierarchical* organization schemes are adopted.[1] Roughly speaking, sub-networks are seen as single host nodes connected to interface nodes called gateways. Gateways perform fairly sophisticated network layer tasks, including routing. Groups of gateways, connected by an arbitrary topology, define logical areas. Inside each area, all the gateways are at the same hierarchical level and "flat" routing is performed among them. Areas communicate only by means of area border gateways. In this way, the computational complexity of the routing problem, as seen by each gateway, is much reduced (e.g., in the Internet, OSPF areas typically group 10 to 300 gateways), while the complexity of the design and management of the routing protocol is much increased.

The instance of our communication network is mapped on a directed weighted graph with $N$ processing/forwarding nodes. All the links are viewed as bit pipes characterized by a bandwidth (bit/sec) and a transmission delay (sec), and are accessed following a statistical multiplexing scheme. For this purpose, every node, of type store-and-forward, holds a buffer space where the incoming and the outgoing packets are stored. This buffer is a shared resource among all the queues attached to every incoming and outgoing link of the node. All the traveling packets are subdivided in two classes: data and routing packets. All the packets in the same class have the same priority, so they are queued and served on the basis of a first-in-first-out policy, but routing packets have a greater priority than data packets. The workload is defined in terms of applications whose arrival rate is dictated by a selected probabilistic model. By application (or session, or connection in the following), we mean a process sending data packets from an origin node to a destination node. The number of packets to send, their sizes and the intervals between them are assigned according to some defined stochastic process. We didn't make any distinction among nodes, they act at the same time as hosts (session end-points) and gateways/routers (forwarding elements). The adopted workload model incorporates a simple flow control mechanism implemented by using a fixed production window for the session's packets generation. The window determines the maximum number of data packets waiting to be sent. Once sent, a packet is considered to be acknowledged. This means that the transport layer neither manages error control, nor packet sequencing, nor acknowledgements and retransmissions.[2]

For each incoming packet, the node's routing component uses the information stored in the local routing table to assign the outgoing link to be used to forward the packet toward its target node. When the link resources are available, they are reserved and the transfer is set up. The time it takes to move a packet from one node to a neighboring one depends on the packet size and on the link transmission characteristics. If, on a packet's arrival, there is not enough buffer space to hold it, the packet is discarded. Otherwise, a service time is stochastically generated for the newly arrived packet. This time represents the delay between the packet arrival time and the time when it will be put in the buffer queue of the outgoing link the local routing component has selected for it.

Situations causing a temporary or steady alteration of the network topology or of its physical characteristics are not taken into account (link or node failure, adding or deleting of network components, etc.).

---

1. A hierarchical structure is adopted on the Internet, organized in hierarchical Autonomous Systems and multiple routing areas inside each Autonomous System (Moy, 1998).
2. This choice is the same as in the "Simple_Traffic" model in the MaRS network simulator (Alaettinoğlu, Shankar, Dussa-Zieger, & Matta, 1992). It can be seen as a very basic form of File Transfer Protocol (FTP).

We developed a complete network simulator in C++. It is a discrete event simulator using as its main data structure an event list, which holds the next future events. The simulation time is a continuous variable and is set by the currently scheduled event. The aim of the simulator is to closely mirror the essential features of the concurrent and distributed behavior of a generic communication network without sacrificing efficiency and flexibility in code development.

We end this section with some remarks concerning two features of the model.

First, we chose not to implement a "real" transport layer for a proper management of error, flow, and congestion control. In fact, each additional control component has a considerable impact on the network performance,[3] making very difficult to evaluate and to study the properties of each control algorithm without taking in consideration the complex way it interacts with all the other control components. Therefore, we chose to test the behavior of our algorithm and of its competitors in conditions such that the number of interacting components is minimal and the routing component can be evaluated in isolation, allowing a better understanding of its properties. To study routing in conjunction with error, flow and congestion control, all these components should be designed at the same time, to allow a good match among their characteristics to produce a synergetic effect.

Second, we chose to work with connection-less and not with connection-oriented networks because connection-oriented schemes are mainly used in networks able to deliver Quality of Service (QoS) (Crawley, Nair, Rajagopalan, & Sandick, 1996).[4] In this case, suitable admission control algorithms have to be introduced, taking into account many economic and technological factors (Sandick & Crawley, 1997). But, again, as a first step we think that it is more reasonable to try to check the validity of a routing algorithm by reducing the number of components heavily influencing the network behavior.

## 4. AntNet: An Adaptive Agent-based Routing Algorithm

The characteristics of the routing problem (discussed in Section 2.2) make it well suited to be solved by a mobile multi-agent approach (Stone & Veloso, 1996; Gray, Kotz, Nog, Rus, & Cybenko, 1997). This processing paradigm is a good match for the distributed and non-stationary (in topology and traffic patterns) nature of the problem, presents a high level of redundancy and fault-tolerance, and can handle multiple objectives and constraints in a flexible way.

*AntNet*, the routing algorithm we propose in this paper, is a mobile agents system showing some essential features of parallel replicated Monte Carlo systems (Streltsov & Vakili, 1996). AntNet takes inspiration from previous work on artificial ant colonies techniques to solve combinatorial optimization problems (Dorigo et al., 1991; Dorigo, 1992; Dorigo et al., 1996; Dorigo & Gambardella, 1997) and telephone network routing (Schoonderwoerd et al., 1996, 1997). The core ideas of these techniques (for a review see Dorigo, Di Caro, and Gambardella, 1998) are (i) the use of repeated and concurrent simulations carried out by a population of artificial agents called "ants" to generate new solutions to the problem, (ii) the use by the agents of stochastic local search to build the solutions in an incremental way, and (iii) the use of information collected during past simulations to direct future search for better solutions.

---

3. As an example, some authors reported an improvement ranging from 2 to 30% in various performance measures for real Internet traffic (Danzig, Liu, & Yan, 1994) by changing from the Reno version to the Vegas version of the TCP (Peterson & Davie, 1996) (the current Internet Transport Control Protocol), and other authors even claimed improvements ranging from 40 to 70% (Brakmo, O'Malley, & Peterson, 1994).

4. This is not the case for the current Internet, where the IP bearer service is of "best-effort" type, meaning that it does the best it can but no guarantees of service quality in terms of delay or bandwidth or jitter, etc., can be assured.

In the artificial ant colony approach, following an iterative process, each ant builds a solution by using two types of information locally accessible: problem-specific information (for example, distance among cities in a traveling salesman problem), and information added by ants during previous iterations of the algorithm. In fact, while building a solution, each ant collects information on the problem characteristics and on its own performance, and uses this information to modify the representation of the problem, as seen locally by the other ants. The representation of the problem is modified in such a way that information contained in past good solutions can be exploited to build new better solutions. This form of indirect communication mediated by the environment is called *stigmergy*, and is typical of social insects (Grassé, 1959).

In AntNet, we retain the core ideas of the artificial ant colony paradigm, and we apply them to solve in an adaptive way the routing problem in datagram networks.

Informally, the AntNet algorithm and its main characteristics can be summarized as follows.

- At regular intervals, and concurrently with the data traffic, from each network node mobile agents are asynchronously launched towards randomly selected destination nodes.

- Agents act concurrently and independently, and communicate in an indirect way, through the information they read and write locally to the nodes.

- Each agent searches for a minimum cost path joining its source and destination nodes.

- Each agent moves step-by-step towards its destination node. At each intermediate node a greedy stochastic policy is applied to choose the next node to move to. The policy makes use of (i) local agent-generated and maintained information, (ii) local problem-dependent heuristic information, and (iii) agent-private information.

- While moving, the agents collect information about the time length, the congestion status and the node identifiers of the followed path.

- Once they have arrived at the destination, the agents go back to their source nodes by moving along the same path as before but in the opposite direction.

- During this backward travel, local models of the network status and the local routing table of each visited node are modified by the agents as a function of the path they followed and of its goodness.

- Once they have returned to their source node, the agents die.

In the following subsections the above scheme is explained, all its components are explicated and discussed, and a more detailed description of the algorithm is given.

## 4.1 Algorithm Description and Characteristics

AntNet is conveniently described in terms of two sets of *homogeneous mobile agents* (Stone & Veloso, 1996), called in the following *forward* and *backward ants*. Agents[5] in each set possess the same structure, but they are differently situated in the environment; that is, they can sense different inputs and they can produce different, independent outputs. They can be broadly classified as *deliberative* agents, because they behave *reactively* retrieving a pre-compiled set of behaviors, and at the same time they maintain a complete internal state description. Agents communicate in an indirect way, according to the stigmergy paradigm, through the information they concurrently read and write in two data structures stored in each network node $k$ (see Figure 1):

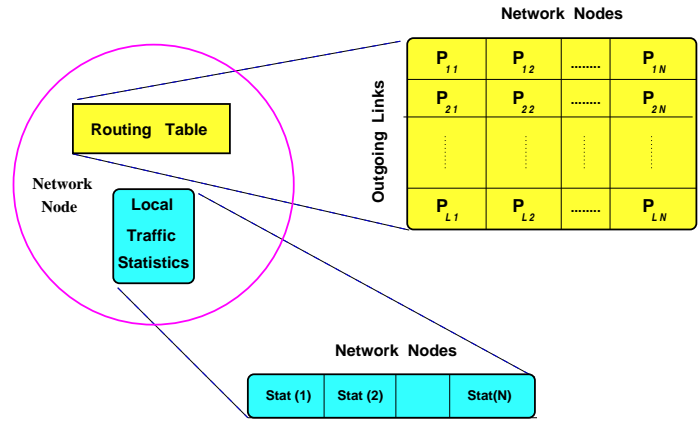

Figure 1: Node structures used by mobile agents in AntNet for the case of a node with $L$ neighbors and a network with $N$ nodes. The routing table is organized as in vector-distance algorithms, but the entries are probabilistic values. The structure containing statistics about the local traffic plays the role of a local adaptive model for the traffic toward each possible destination.

i) A routing table $\mathcal{T}_k$, organized as in vector-distance algorithms (see Appendix A), but with probabilistic entries. $\mathcal{T}_k$ defines the probabilistic routing policy currently adopted at node $k$: for each possible destination $d$ and for each neighbor node $n$, $\mathcal{T}_k$ stores a probability value $P_{nd}$ expressing the goodness (desirability), under the current network-wide routing policy, of choosing $n$ as next node when the destination node is $d$:

$$\sum_{n \in \mathcal{N}_k} P_{nd} = 1, \quad d \in [1, N], \quad \mathcal{N}_k = \{neighbors(k)\}.$$

ii) An array $\mathcal{M}_k(\mu_d, \sigma_d^2, \mathcal{W}_d)$, of data structures defining a simple parametric statistical model for the traffic distribution over the network as seen by the local node $k$. The model is adaptive and described by sample means and variances computed over the trip times experienced by the mobile agents, and by a moving observation window $\mathcal{W}_d$ used to store the best value $W_{best_d}$ of the agents' trip time.

---

5. In the following, we will use interchangeably the terms ant and agent.

For each destination $d$ in the network, an estimated mean and variance, $\mu_d$ and $\sigma_d{}^2$, give a representation of the expected time to go and of its stability. We used arithmetic, exponential and windowed strategies to compute the statistics. Changing strategy does not affect performance much, but we observed the best results using the exponential model:[6]

$$
\begin{aligned}
\mu_d &\leftarrow \mu_d + \eta(o_{k \to d} - \mu_d), \\
\sigma_d{}^2 &\leftarrow \sigma_d{}^2 + \eta((o_{k \to d} - \mu_d)^2 - \sigma_d{}^2),
\end{aligned}
\tag{1}
$$

where $o_{k \to d}$ is the new observed agent's trip time from node $k$ to destination $d$.[7]

The moving observation window $\mathcal{W}_d$ is used to compute the value $W_{best_d}$ of the best agents' trip time towards destination $d$ as observed in the last $w$ samples. After each new sample, $w$ is incremented modulus $|\mathcal{W}|_{max}$, and $|\mathcal{W}|_{max}$ is the maximum allowed size of the observation window. The value $W_{best_d}$ represents a short-term memory expressing a moving empirical lower bound of the estimate of the time to go to node $d$ from the current node.

$\mathcal{T}$ and $\mathcal{M}$ can be seen as memories local to nodes capturing different aspects of the network dynamics. The model $\mathcal{M}$ maintains absolute distance/time estimates to all the nodes, while the routing table gives relative probabilistic goodness measures for each link-destination pair under the current routing policy implemented over all the network.

The AntNet algorithm is described as follows.

1. At regular intervals $\Delta t$ from every network node $s$, a mobile agent (forward ant) $F_{s \to d}$ is launched toward a destination node $d$ to discover a feasible, low-cost path to that node and to investigate the load status of the network. Forward ants share the same queues as data packets, so that they experience the same traffic loads. Destinations are locally selected according to the data traffic patterns generated by the local workload: if $f_{sd}$ is a measure (in bits or in number of packets) of the data flow $s \to d$, then the probability of creating at node $s$ a forward ant with node $d$ as destination is

$$
p_d = \frac{f_{sd}}{\displaystyle\sum_{d'=1}^{N} f_{sd'}}.
\tag{2}
$$

In this way, ants adapt their exploration activity to the varying data traffic distribution.

2. While traveling toward their destination nodes, the agents keep memory of their paths and of the traffic conditions found. The identifier of every visited node $k$ and the time elapsed since the launching time to arrive at this $k$-th node are pushed onto a memory stack $S_{s \to d}(k)$.

---

6. This is the same model as used by the Jacobson/Karels algorithm to estimate retransmission timeouts in the Internet TCP(Peterson & Davie, 1996).

7. The factor $\eta$ weights the number of most recent samples that will really affect the average. The weight of the $t_i$-th sample used to estimate the value of $\mu_d$ after $j$ samplings, with $j > i$, is: $\eta(1-\eta)^{j-i}$. In this way, for example, if $\eta = 0.1$, approximately only the latest 50 observations will really influence the estimate, for $\eta = 0.05$, the latest 100, and so on. Therefore, the number of effective observations is $\approx 5(1/\eta)$.

3. At each node $k$, each traveling agent headed towards its destination $d$ selects the node $n$ to move to choosing among the neighbors it did not already visit, or over all the neighbors in case all of them had been previously visited. The neighbor $n$ is selected with a probability (goodness) $P'_{nd}$ computed as the normalized sum of the probabilistic entry $P_{nd}$ of the routing table with a heuristic correction factor $l_n$ taking into account the state (the length) of the $n$-th link queue of the current node $k$:

$$P'_{nd} = \frac{P_{nd} + \alpha l_n}{1 + \alpha(|\mathcal{N}_k| - 1)}. \tag{3}$$

The heuristic correction $l_n$ is a [0,1] normalized value proportional to the length $q_n$ (in bits waiting to be sent) of the queue of the link connecting the node $k$ with its neighbor $n$:

$$l_n = 1 - \frac{q_n}{\sum\limits_{n'=1}^{|\mathcal{N}_k|} q_{n'}}. \tag{4}$$

The value of $\alpha$ weights the importance of the heuristic correction with respect to the probability values stored in the routing table. $l_n$ reflects the instantaneous state of the node's queues, and assuming that the queue's consuming process is almost stationary or slowly varying, $l_n$ gives a quantitative measure associated with the queue waiting time. The routing tables values, on the other hand, are the outcome of a continual learning process and capture both the current and the past status of the whole network as seen by the local node. Correcting these values with the values of $l$ allows the system to be more "reactive", at the same time avoiding following all the network fluctuations. Agent's decisions are taken on the basis of a combination of a long-term learning process and an instantaneous heuristic prediction.

In all the experiments we ran, we observed that the introduced correction is a very effective mechanism. Depending on the characteristics of the problem, the best value to assign to the weight $\alpha$ can vary, but if $\alpha$ ranges between 0.2 and 0.5, performance doesn't change appreciably. For lower values, the effect of $l$ is vanishing, while for higher values the resulting routing tables oscillate and, in both cases, performance degrades.

4. If a cycle is detected, that is, if an ant is forced to return to an already visited node, the cycle's nodes are popped from the ant's stack and all the memory about them is destroyed. If the cycle lasted longer than the lifetime of the ant before entering the cycle, (that is, if the cycle is greater than half the ant's age) the ant is destroyed. In fact, in this case the agent wasted a lot of time probably because of a wrong sequence of decisions and not because of congestion states. Therefore, the agent is carrying an old and misleading memory of the network state and it is counterproductive to use it to update the routing tables (see below).

5. When the destination node $d$ is reached, the agent $F_{s \to d}$ generates another agent (backward ant) $B_{d \to s}$, transfers to it all of its memory, and dies.

6. The backward ant takes the same path as that of its corresponding forward ant, but in the opposite direction.[8] At each node $k$ along the path it pops its stack $S_{s \to d}(k)$ to know the next hop node. Backward ants do not share the same link queues as data packets; they use higher priority queues, because their task is to quickly propagate to the routing tables the information accumulated by the forward ants.

7. Arriving at a node $k$ coming from a neighbor node $f$, the backward ant updates the two main data structures of the node, the local model of the traffic $\mathcal{M}_k$ and the routing table $\mathcal{T}_k$, for all the entries corresponding to the (forward ant) destination node $d$. With some precautions, updates are performed also on the entries corresponding to every node $k' \in S_{k \to d}, k' \neq d$ on the "sub-paths" followed by ant $F_{s \to d}$ after visiting the current node $k$. In fact, if the elapsed trip time of a sub-path is statistically "good" (i.e., it is less than $\mu + I(\mu, \sigma)$, where $I$ is an estimate of a confidence interval for $\mu$), then the time value is used to update the corresponding statistics and the routing table. On the contrary, trip times of sub-paths not deemed good, in the same statistical sense as defined above, are not used because they don't give a correct idea of the time to go toward the sub-destination node. In fact, all the forward ant routing decisions were made only as a function of the destination node. In this perspective, sub-paths are side effects, and they are intrinsically sub-optimal because of the local variations in the traffic load (we can't reason with the same perspective as in dynamic programming, because of the non-stationarity of the problem representation). Obviously, in case of a good sub-path we can use it: the ant discovered, at zero cost, an additional good route. In the following two items the way $\mathcal{M}$ and $\mathcal{T}$ are updated is described with respect to a generic "destination" node $d' \in S_{k \to d}$.

   i) $\mathcal{M}_k$ is updated with the values stored in the stack memory $S_{s \to d}(k)$. The time elapsed to arrive (for the forward ant) to the destination node $d'$ starting from the current node is used to update the mean and variance estimates, $\mu_{d'}$ and $\sigma_{d'}{}^2$, and the best value over the observation window $\mathcal{W}_{d'}$. In this way, a parametric model of the traveling time to destination $d'$ is maintained. The mean value of this time and its dispersion can vary strongly, depending on the traffic conditions: a poor time (path) under low traffic load can be a very good one under heavy traffic load. The statistical model has to be able to capture this variability and to follow in a robust way the fluctuations of the traffic. This model plays a critical role in the routing table updating process (see item (ii) below). Therefore, we investigated several ways to build effective and computationally inexpensive models, as described in the following Section 4.2.

   ii) The routing table $\mathcal{T}_k$ is changed by incrementing the probability $P_{fd'}$ (i.e., the probability of choosing neighbor $f$ when destination is $d'$) and decrementing, by normalization, the other probabilities $P_{nd'}$. The amount of the variation in the probabilities depends on a measure of goodness we associate with the trip time $T_{k \to d'}$ experienced by the forward ant, and is given below. This time represents the only available explicit feedback signal to score paths. It gives a clear indication about the goodness $r$ of the followed route because it is proportional to its length from a physical point of view (number of hops, transmission capacity of the used links, processing speed of the crossed nodes) and from a traffic congestion point of view (the forward ants share the same queues as data packets).

---

8. This assumption requires that all the links in the network are bi-directional. In modern networks this is a reasonable assumption.

The time measure $T$, composed by all the sub-paths elapsed times, cannot be associated with an exact error measure, given that we don't know the "optimal" trip times, which depend on the whole network load status.[9] Therefore, $T$ can only be used as a reinforcement signal. This gives rise to a credit assignment problem typical of the reinforcement learning field (Bertsekas & Tsitsiklis, 1996; Kaelbling et al., 1996). We define the reinforcement $r \equiv r(T, \mathcal{M}_k)$ to be a function of the goodness of the observed trip time as estimated on the basis of the local traffic model. $r$ is a dimensionless value, $r \in (0, 1]$, used by the current node $k$ as a positive reinforcement for the node $f$ the backward ant $B_{d \to s}$ comes from. $r$ takes into account some average of the so far observed values and of their dispersion to score the goodness of the trip time $T$, such that the smaller $T$ is, the higher $r$ is (the exact definition of $r$ is discussed in the next subsection). The probability $P_{fd'}$ is increased by the reinforcement value as follows:

$$P_{fd'} \leftarrow P_{fd'} + r(1 - P_{fd'}).  \qquad (5)$$

In this way, the probability $P_{fd'}$ will be increased by a value proportional to the reinforcement received and to the previous value of the node probability (that is, given a same reinforcement, small probability values are increased proportionally more than big probability values, favoring in this way a quick exploitation of new, and good, discovered paths).

Probabilities $P_{nd'}$ for destination $d'$ of the other neighboring nodes $n$ implicitly receive a negative reinforcement by normalization. That is, their values are reduced so that the sum of probabilities will still be 1:

$$P_{nd'} \leftarrow P_{nd'} - rP_{nd'}, \quad n \in \mathcal{N}_k, \ n \neq f.  \qquad (6)$$

It is important to remark that every discovered path receives a positive reinforcement in its selection probability, and the reinforcement is (in general) a non-linear function of the goodness of the path, as estimated using the associated trip time. In this way, not only the (explicit) assigned value $r$ plays a role, but also the (implicit) ant's arrival rate. This strategy is based on trusting paths that receive either high reinforcements, independent of their frequency, or low and frequent reinforcements. In fact, for any traffic load condition, a path receives one or more high reinforcements only if it is much better than previously explored paths. On the other hand, during a transient phase after a sudden increase in network load all paths will likely have high traversing times with respect to those learned by the model $\mathcal{M}$ in the preceding, low congestion, situation. Therefore, in this case good paths can only be differentiated by the frequency of ants' arrivals.

---

9. When the network is in a congested state, all the trip times will score poorly with respect to the times observed in low load situations. Nevertheless, a path with a high trip time should be scored as a good path if its trip time is significantly lower than the other trip times observed in the same congested situation.

Assigning always a positive, but low, reinforcement value in the case of paths with high traversal time allows the implementation of the above mechanism based on the frequency of the reinforcements, while, at the same time, avoids giving excessive credit to paths with high traversal time due to their poor quality.

The use of probabilistic entries is very specific to AntNet and we observed it to be effective, improving the performance, in some cases, even by 30%-40%. Routing tables are used in a probabilistic way not only by the ants but also by the data packets. This has been observed to improve AntNet performance, which means that the way the routing tables are built in AntNet is well matched with a probabilistic distribution of the data packets over all the good paths. Data packets are prevented from choosing links with very low probability by re-mapping the $\mathcal{T}$'s entries by means of a power function $f(p) = p^\alpha, \alpha > 1$, which emphasizes high probability values and reduces lower ones (in our experiments we set $\alpha$ to 1.2).

Figure 2 gives a high-level description of the algorithm in pseudo-code, while Figure 3 illustrates a simple example of the algorithm behavior. A detailed discussion of the characteristics of the algorithm is postponed to Section 8, after the performance of the algorithm has been analyzed with respect to a set of competitor algorithms. In this way, the characteristics of AntNet can be meaningfully evaluated and compared to those of other state-of-the-art algorithms.

## 4.2 How to Score the Goodness of the Ant's Trip Time

The reinforcement $r$ is a critical quantity that has to be assigned by considering three main aspects: (i) paths should receive an increment in their selection probability proportional to their goodness, (ii) the goodness is a relative measure, which depends on the traffic conditions, that can be estimated by means of the model $\mathcal{M}$, and (iii) it is important not to follow all the traffic fluctuations. This last aspect is particularly important. Uncontrolled oscillations in the routing tables are one of the main problems in shortest paths routing (Wang & Crowcroft, 1992). It is very important to be able to set the best trade-off between stability and adaptivity.

We investigated several ways to assign the $r$ values trying to take into account the above three requirements:

- The simplest way is to set $r = constant$: independently of the ant's "experiment outcomes", the discovered paths are all rewarded in the same way. In this simple but meaningful case, what is at work is the implicit reinforcement mechanism due to the differentiation in the ant arrival rates. Ants traveling along faster paths will arrive at a higher rate than other ants, hence their paths will receive a higher cumulative reward.[10] The obvious problem of this approach lies in the fact that, although ants following longer paths arrive delayed, they will nevertheless have the same effect on the routing tables as the ants who followed shorter paths.

---

10. In this case, the core of the algorithm is based on the capability of "real" ants to discover shortest paths communicating by means of pheromone trails (Goss et al., 1989; Beckers et al., 1992).

```
t     :=  Current time;
t_end :=  Time length of the simulation;
Δt    :=  Time interval between ants generation;
foreach (Node)   /∗ Concurrent activity over the network ∗/
   M = Local traffic model;
   T = Node routing table;
   while ( t ≤ t_end )
      in_parallel   /∗ Concurrent activity on each node ∗/
         if ( t mod Δt = 0)
            destination_node := SelectDestinationNode(data_traffic_distribution);
            LaunchForwardAnt(destination_node, source_node);
         end if
         foreach (ActiveForwardAnt[source_node, current_node, destination_node])
            while (current_node ≠ destination_node)
               next_hop_node := SelectLink(current_node, destination_node, T, link_queues);
               PutAntOnLinkQueue(current_node, next_hop_node);
               WaitOnDataLinkQueue(current_node, next_hop_node);
               CrossTheLink(current_node, next_hop_node);
               PushOnTheStack(next_hop_node, elapsed_time);
               current_node := next_hop_node;
            end while
            LaunchBackwardAnt(destination_node, source_node, stack_data);
            Die();
         end foreach
         foreach (ActiveBackwardAnt[source_node, current_node, destination_node])
            while (current_node ≠ destination_node)
               next_hop_node := PopTheStack();
               WaitOnHighPriorityLinkQueue(current_node, next_hop_node);
               CrossTheLink(current_node, next_hop_node);
               UpdateLocalTrafficModel(M, current_node, source_node, stack_data);
               reinforcement := GetReinforcement(current_node, source_node, stack_data, M);
               UpdateLocalRoutingTable(T, current_node, source_node, reinforcement);
            end while
         end foreach
      end in_parallel
   end while
end foreach
```

Figure 2: AntNet's top-level description in pseudo-code. All the described actions take place in a completely distributed and concurrent way over the network nodes (while, in the text, AntNet has been described from an individual ant's perspective). All the constructs at the same level of indentation inside the context of the statement **in_parallel** are executed concurrently. The processes of data generation and forwarding are not described, but they can be thought as acting concurrently with the ants.

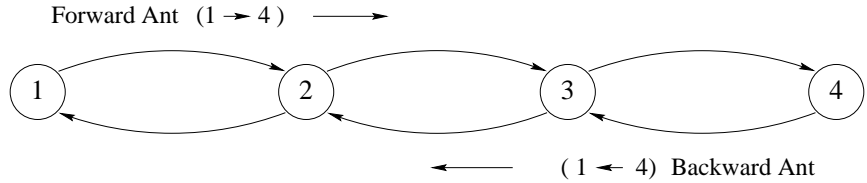

Figure 3: Example of AntNet behavior. The forward ant, $F_{1\to4}$, moves along the path $1 \to 2 \to 3 \to 4$ and, arrived at node 4, launches the backward ant $B_{4\to1}$ that will travel in the opposite direction. At each node $k$, $k = 3, \ldots, 1$, the backward ant will use the stack contents $S_{1\to4}(k)$ to update the values for $\mathcal{M}_k(\mu_4, \sigma_4{}^2, \mathcal{W}_4)$, and, in case of good sub-paths, to update also the values for $\mathcal{M}_k(\mu_i, \sigma_i{}^2, \mathcal{W}_i)$, $i = k+1, \ldots, 3$. At the same time the routing table will be updated by incrementing the goodness $P_{j4}$, $j = k+1$, of the last node $k+1$ the ant $B_{4\to1}$ came from, for the case of node $i = k+1, \ldots, 4$ as destination node, and decrementing the values of $P$ for the other neighbors (here not shown). The increment will be a function of the trip time experienced by the forward ant going from node $k$ to destination node $i$. As for $\mathcal{M}$, the routing table is always updated for the case of node 4 as destination, while the other nodes $i' = k+1, \ldots, 3$ on the sub-paths are taken in consideration as destination nodes only if the trip time associated to the corresponding sub-path of the forward ant is statistically good.

In the experiments we ran with this strategy, the algorithm showed moderately good performance. These results suggest that the "implicit" component of the algorithm, based on the ant arrival rate, plays a very important role. Of course, to compete with state-of-the-art algorithms, the available information about path costs has to be used.

- More elaborate approaches define $r$ as a function of the ant's trip time $T$, and of the parameters of the local statistical model $\mathcal{M}$. We tested several alternatives, by using different linear, quadratic and hyperbolic combinations of the $T$ and $\mathcal{M}$ values. In the following we limit the discussion to the functional form that gave the best results, and that we used in the reported experiments:

$$r = c_1 \left( \frac{W_{best}}{T} \right) + c_2 \left( \frac{I_{sup} - I_{inf}}{(I_{sup} - I_{inf}) + (T - I_{inf})} \right). \tag{7}$$

In Equation 7, $W_{best}$ is the best trip time experienced by the ants traveling toward the destination $d$, over the last observation window $\mathcal{W}$. The maximum size of the window (the maximum number of considered samples before resetting the $W_{best}$ value) is assigned on the basis of the coefficient $\eta$ of Equation 1. As we said, $\eta$ weights the number of samples effectively giving a contribution to the value of the $\mu$ estimate, defining a sort of moving exponential window. Following the expression for the number of effective samples as reported in footnote 7, we set $|\mathcal{W}|_{max} = 5(c/\eta)$, with $c < 1$. In this way, the long-term exponential mean and the short-term windowing are referring to a comparable set of observations, with the short-term mean evaluated over a fraction $c$ of the samples used for the long-term one. $I_{sup}$ and $I_{inf}$ are convenient estimates of the limits of an approximate confidence interval for $\mu$. $I_{inf}$ is set to $W_{best}$, while $I_{sup} = \mu + z(\sigma/\sqrt{|\mathcal{W}|})$, with $z = 1/\sqrt{(1-\gamma)}$ where $\gamma$ gives the selected confidence level.[11] There is some level of arbitrariness in our computation of the confidence interval, because we set it in an asymmetric way and $\mu$ and $\sigma$ are not arithmetic estimates. Anyway, what we need is a quick, raw estimate of the mean value and of the dispersion of the values (for example, a local bootstrap procedure could have been applied to extract a meaningful confidence interval, but such a choice is not reasonable from a CPU time-consuming perspective).

The first term in Equation 7 simply evaluates the ratio between the current trip time and the best trip time observed over the current observation window. This term is corrected by the second one, that evaluates how far the value $T$ is from $I_{inf}$ in relation to the extension of the confidence interval, that is, considering the stability in the latest trip times. The coefficients $c_1$ and $c_2$ weight the importance of each term. The first term is the most important one, while the second term plays the role of a correction. In the current implementation of the algorithm we set $c_1 = 0.7$ and $c_2 = 0.3$. We observed that $c_2$ shouldn't be too big (0.35 is an upper limit), otherwise performance starts to degrade appreciably. The behavior of the algorithm is quite stable for $c_2$ values in the range 0.15 to 0.35 but setting $c_2$ below 0.15 slightly degrades performance. The algorithm is very robust to changes in $\gamma$, which defines the confidence level: varying the confidence level in the range from 75% to 95% changes performance little. The best results have been obtained for values around 75%÷80%. We observed that the algorithm is very robust to its internal parameter settings and we didn't try to "adapt" the set of parameters to the problem instance. All the different experiments were carried out with the same "reasonable" settings. We could surely improve the performance by means of a finer tuning of the parameters, but we didn't because we were interested in implementing a robust system, considering that the world of networks is incredibly varied in terms of traffic, topologies, switch and transmission characteristics, etc.

The value $r$ obtained from Equation 7 is finally transformed by means of a squash function $s(x)$:

$$s(x) = \left(1 + exp\left(\frac{a}{x|\mathcal{N}_k|}\right)\right)^{-1}, \qquad x \in (0,1], \quad a \in R^+,  \tag{8}$$

$$r \leftarrow \frac{s(r)}{s(1)}.  \tag{9}$$

Squashing the $r$ values allows the system to be more sensitive in rewarding good (high) values of $r$, while having the tendency to saturate the rewards for bad (near to zero) $r$ values: the scale is compressed for lower values and expanded in the upper part. In such a way an emphasis is put on good results, while bad results play a minor role.

---

11. The expression is obtained by using the Tchebycheff inequality that allows the definition of a confidence interval for a random variable following any distribution (Papoulis, 1991) Usually, for specific probability densities the Tchebycheff bound is too high, but here we can conveniently use it because (i) we want to avoid to make assumptions on the distribution of $\mu$ and, (ii) we need only a raw estimate of the confidence interval.

The coefficient $a/|\mathcal{N}_k|$ determines a parametric dependence of the squashed reinforcement value on the number $|\mathcal{N}_k|$ of neighbors of the reinforced node $k$: the greater the number of neighbors, the higher the reinforcement (see Figure 4). The reason to do this is that we want to have a similar, strong, effect of good results on the probabilistic routing tables, independent of the number of neighbor nodes.

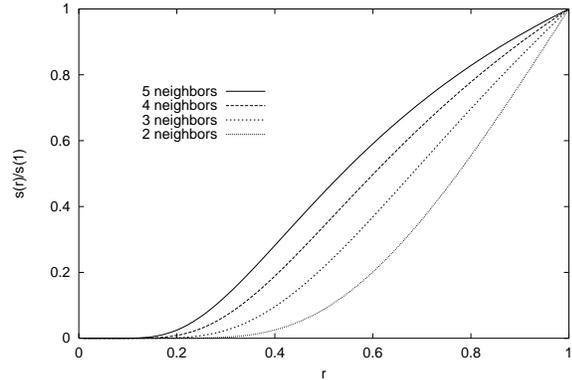

Figure 4: Examples of squash functions with a variable number of node neighbors.

# 5. Routing Algorithms Used for Comparison

To evaluate the performance of AntNet, we compared it with state-of-the-art routing algorithms from the telecommunications and machine learning fields. The following algorithms, belonging to the various possible combinations of static and adaptive, distance-vector and link-state classes (see Appendix A), have been implemented and used to run comparisons.

**OSPF (static, link state):** is our implementation of the current Interior Gateway Protocol (IGP) of Internet (Moy, 1998). Being interested in studying routing under the assumptions described in Section 3, the routing protocol we implemented does not mirror the real OSPF protocol in all its details. It only retains the basic features of OSPF. Link costs are statically assigned on the basis of their physical characteristics and routing tables are set as the result of the shortest (minimum time) path computation for a sample data packet of size 512 bytes. It is worth remarking that this choice penalizes our version of OSPF with respect to the real one. In fact, in the real Internet link costs are set by network administrators who can use additional heuristic and on-field knowledge they have about traffic workloads.

**SPF (adaptive, link-state):** is the prototype of link-state algorithms with dynamic metric for link costs evaluations. A similar algorithm was implemented in the second version of ARPANET (McQuillan, Richer, & Rosen, 1980) and in its successive revisions (Khanna & Zinky, 1989). Our implementation uses the same flooding algorithm, while link costs are assigned over a discrete scale of 20 values by using the ARPANET hop-normalized-delay metric[12] (Khanna & Zinky, 1989) and the the statistical window average method described in (Shankar, Alaettinoğlu, Dussa-Zieger, & Matta, 1992a). Link costs are computed as weighted averages between short and long-term real-valued statistics reflecting the delay (e.g., utilization, queueing and/or transmis-sion delay, etc.) over fixed time intervals. Obtained values are rescaled and saturated by a linear function. We tried several additional discrete and real-valued metrics but the discretized hop-normalized-delay gave the best results in terms of performance and stability. Using a discretized scale reduces the sensitivity of the algorithm but at the same time reduces also undesirable oscillations.

---

12. The transmitting node monitors the average packet delay $\overline{d}$ (queuing and transmission) and the average packet transmission time $\overline{t}$ over fix observation windows. From these measures, assuming an M/M/1 queueing model (Bertsekas & Gallager, 1992), a link utilization cost measure is calculated as $1 - \overline{t}/\overline{d}$.

**BF (adaptive, distance-vector):** is an implementation of the asynchronous distributed Bellman-Ford algorithm with dynamic metrics (Bertsekas & Gallager, 1992; Shankar et al., 1992a). The algorithm has been implemented following the guidelines of Appendix A, while link costs are assigned in the same way as described for SPF above. Vector-distance Bellman-Ford-like algorithms are today in use mainly for intra-domain routing, because they are used in the Routing Information Protocol (RIP) (Malkin & Steenstrup, 1995) supplied with the BSD version of Unix. Several enhanced versions of the basic adaptive Bellman-Ford algorithm can be found in the literature (for example the Merlin-Segall (Merlin & Segall, 1979) and the Extended Bellman-Ford (Cheng, Riley, Kumar, & Garcia-Luna-Aceves, 1989) algorithms). They focus mainly on reducing the information dissemination time in case of link failures. When link failures are not a major issue, as in this paper, their behavior is in general equivalent to that of the basic adaptive Bellman-Ford.

**Q-R (adaptive, distance-vector):** is the Q-Routing algorithm as proposed by Boyan and Littman (1994). This is an online asynchronous version of the Bellman-Ford algorithm. Q-R learns online the values $Q_k(d, n)$, which are estimates of the time to reach node $d$ from node $k$ via the neighbor node $n$. Upon sending a packet $P$ from $k$ to neighbor node $n$ with destination $d$, a back packet $P_{back}$ is immediately generated from $n$ to $k$. $P_{back}$ carries the information about the current time estimate $t_{n \to d} = \min_{n' \in \mathcal{N}_n} Q_n(d, n')$ held at node $n$ about the time to go for destination $d$, and the sum $t_{P_{k \to n}}$ of the queuing and transmission time experienced by $P$ since its arrival at node $k$. The sum $Q_{new}(d, n) = t_{n \to d} + t_{P_{k \to n}}$ is used to compute the variation $\Delta Q_k(d, n) = \eta(Q_{new}(d, n) - Q_k(d, n))$ of the Q-learning-like value $Q_k(d, n)$.

**PQ-R (adaptive, distance-vector):** is the Predictive Q-Routing algorithm (Choi & Yeung, 1996), an extension of Q-Routing. In Q-routing the best link (i.e., the one with the lowest $Q_k(d, n)$) is deterministically chosen by packets. Therefore, a link that happens to have a high expected $Q_k(d, n)$, for example because of a temporary load condition, will never be used again until all the other links exiting from the same node have a worse, that is higher, $Q_k(d, n)$. PQ-R learns a model of the rate of variation of links' queues, called the recovery rate, and uses it to probe those links that, although not having the lowest $Q_k(d, n)$, have a high recovery rate.

**Daemon (adaptive, optimal routing):** is an approximation of an ideal algorithm. It defines an empirical bound on the achievable performance. It gives some information about how much improvement is still possible. In the absence of any *a priori* assumption on traffic statistics, the empirical bound can be defined by an algorithm possessing a "daemon" able to read in every instant the state of all the queues in the network and then calculating instantaneous "real" costs for all the links and assigning paths on the basis of a network-wide shortest paths re-calculation for every packet hop. Links costs used in shortest paths calculations are the following:

$$C_l = d_l + \frac{S_p}{b_l} + (1 - \alpha)\frac{S_{Q(l)}}{b_l} + \alpha\,\frac{\bar{S}_{Q(l)}}{b_l},$$

where $d_l$ is the transmission delay for link $l$, $b_l$ is its bandwidth, $S_p$ is the size (in bits) of the data packet doing the hop, $S_{Q(l)}$ is the size (in bits) of the queue of link $l$, $\bar{S}_{Q(l)}$ is the exponential mean of the size of links queue and it is a correction to the actual size of the link queue on the basis of what observed until that moment. This correction is weighted by the $\alpha$ value set to 0.4. Of course, given the arbitrariness we introduced in calculating $C_l$, it could be possible to define an even better Daemon algorithm.

## 6. Experimental Settings

The functioning of a communication network is governed by many components, which may interact in nonlinear and unpredictable ways. Therefore, the choice of a meaningful testbed to compare competing algorithms is no easy task.

A limited set of classes of tunable components is defined and for each class our choices are explained.

### 6.1 Topology and physical properties of the net

Topology can be defined on the basis of a real net instance or it can defined by hand, to better analyze the influence of important topological features (like diameter, connectivity, etc.).

Nodes are mainly characterized by their buffering and processing capacity, whereas links are characterized by their propagation delay, bandwidth and streams multiplexing scheme. For both, fault probability distributions should be defined.

In our experiments, we used three significant net instances with increasing numbers of nodes. For all of them we describe the main characteristics and we summarize the topological properties by means of a triple of numbers ($\mu$, $\sigma$, $N$) indicating respectively the mean shortest path distance, in terms of hops, between all pairs of nodes, the variance of this average, and the total number of nodes. From these three numbers we can get an idea about the degree of connectivity and balancing of the network. The difficulty of the routing problem roughly increases with the value of these numbers.

- *SimpleNet* (1.9, 0.7, 8) is a small network specifically designed to study some aspects of the behavior of the algorithms we compare. Experiments with SimpleNet were designed to closely study how the different algorithms manage to distribute the load on the different possible paths. SimpleNet is composed of 8 nodes and 9 bi-directional links with a bandwidth of 10 Mbit/s and propagation delay of 1 msec. The topology is shown in Figure 5.

- *NSFNET* (2.2, 0.8, 14) is the old USA T1 backbone (1987). NSFNET is a WAN composed of 14 nodes and 21 bi-directional links with a bandwidth of 1.5 Mbit/s. Its topology is shown in Figure 6. Propagation delays range from 4 to 20 msec. NSFNET is a well balanced network.

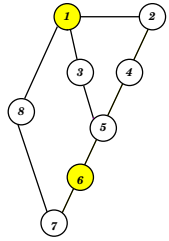

Figure 5: *SimpleNet*. Numbers within circles are node identifiers. Shaded nodes have a special interpretation in our experiments, described later. Each edge in the graph represents a pair of directed links. Link bandwidth is 10 Mbit/sec, propagation delay is 1 msec.

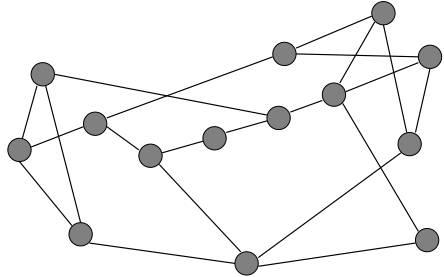

Figure 6: *NSFNET*. Each edge in the graph represents a pair of directed links. Link bandwidth is 1.5 Mbit/sec, propagation delays range from 4 to 20 msec.

- *NTTnet* (6.5, 3.8, 57) is the major Japanese backbone. NTTnet is the NTT (Nippon Telephone and Telegraph company) fiber-optic corporate backbone. NTTnet is a 57 nodes, 162 bi-directional links network. Link bandwidth is of 6 Mbit/sec, while propagation delays range around 1 to 5 msec. The topology is shown in Figure 7. NTTnet is not a well balanced network.

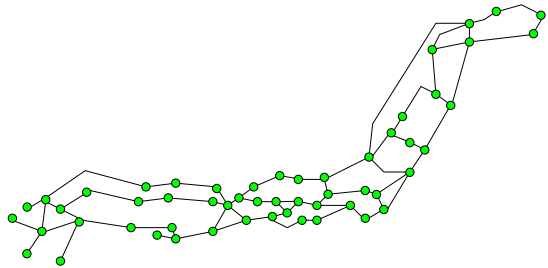

Figure 7: *NTTnet*. Each edge in the graph represents a pair of directed links. Link bandwidth is 6 Mbit/sec, propagation delays range from 1 to 5 msec.

All the networks are simulated with zero link-fault and node-fault probabilities, local node buffers of 1 Gbit capacity, and data packets maximum time to live (TTL) set to 15 sec.

## 6.2 Traffic patterns

Traffic is defined in terms of open sessions between pairs of different nodes. Traffic patterns can show a huge variety of forms, depending on the characteristics of each session and on their distribution from geographical and temporal points of view.

Each single session is characterized by the number of transmitted packets, and by their size and inter-arrival time distributions. More generally, priority, costs and requested quality of service should be used to completely characterize a session.

Sessions over the network can be characterized by their inter-arrival time distribution and by their geographical distribution. The latter is controlled by the probability assigned to each node to be selected as a session start or end-point.

We considered three basic patterns for the temporal distribution of the sessions, and three for their spatial distribution.

Temporal distributions:

- *Poisson* (P): for each node a Poisson process is defined which regulates the arrival of new sessions, i.e., sessions inter-arrival times are negative exponentially distributed.
- *Fixed* (F): at the beginning of the simulation, for each node, a fixed number of one-to-all sessions is set up and left constant for the remainder of the simulation.
- *Temporary* (TMPHS): a temporary, heavy load, traffic condition is generated turning on some nodes that act like hot spots (see below).

Spatial distributions:

- *Uniform* (U): the assigned temporal characteristics for session arrivals are set identically for all the network nodes.
- *Random* (R): in this case, the assigned temporal characteristics for session arrivals are set in a random way over the network nodes.
- *Hot Spots* (HS): some nodes behave as hot spots, concentrating a high rate of input/output traffic. A fixed number of sessions are opened from the hot spots to all the other nodes.

General traffic patterns have been obtained combining the above temporal and spatial characteristics. Therefore, for example, UP traffic means that, for each node, an identical Poisson process is regulating the arrival of new sessions, while in the RP case the process is different for each node, and UP-HS means that a Hot Spots traffic model is superimposed to a UP traffic.

Concerning the shape of the bit stream generated by each session, we consider two basic types:

- *Constant Bit Rate* (CBR): the per-session bit rate is maintained fixed. Examples of applications of CBR streams are the voice signal in a telephone network, which is converted into a stream of bits with a constant rate of 64 Kbit/sec, and the MPEG1 compression standard, which converts a video signal in a stream of 1.5 Mbit/sec.

- *Generic Variable Bit Rate* (GVBR): the per-session generated bit rate is time varying. The term GVBR is a broad generalization of the VBR term normally used to designate a bit stream with a variable bit rate but with known average characteristics and expected/admitted fluctuations.[13] Here, a GVBR session generates packets whose sizes and inter-arrival times are variable and follow a negative exponential distribution. The information about these characteristics is never directly used by the routing algorithms, like in IP-based networks.

The values we used in the experiments to shape traffic patterns are "reasonable" values for session generations and data packet production taking into consideration current network usage and computing power. The mean of the packet size distribution has been set to 4096 bits in all the experiments. Basic temporal and spatial distributions have been chosen to be representative of a wide class of possible situations that can be arbitrarily composed to generate a meaningful subset of real traffic patterns.

### 6.3 Metrics for performance evaluation

Depending on the type of services delivered on the network and on their associated costs, many performance metrics could be defined. We focused on standard metrics for performance evaluation, considering only sessions with equal costs, benefits and priority and without the possibility of requests for special services like real-time. In this framework, the measures we are interested in are: *throughput* (correctly delivered bits/sec), *delay distribution* for data packets (sec), and *network capacity usage* (for data and routing packets), expressed as the sum of the used link capacities divided by the total available link capacity.

### 6.4 Routing algorithms parameters

All the algorithms used have a collection of parameters to be set. Common parameters are routing packet size and elaboration time. Settings for these parameters are shown in table 1. These parameters have been assigned to values used in previous simulation works (Alaettinoğlu et al., 1992) and/or on the basis of heuristic evaluations taking into consideration information encoding schemes and currently available computing power (e.g., the size for forward ants has been determined as the same size of a BF packet plus 8 bytes for each hop to store the information about the node address and the elapsed time). Concerning the other main parameters, specific for each algorithm, for the AntNet competitors we used the best settings we could find in the literature and/or we tried to tune the parameters as much as possible to obtain better results. For OSPF, SPF, and BF, the length of the time interval between consecutive routing information broadcasts and the length of the time window to average link costs are the same, and they are set to 0.8 or 3 seconds, depending on the experiment for SPF and BF, and to 30 seconds for OSPF. Link costs inside each window are assigned as the weighted sum between the arithmetic average over the window and the exponential average with decay factor equal to 0.9. The obtained values are discretized over a linear scale saturated between 1 and 20, with slope set to 20 and maximum admitted variation equal to 1. For Q-R and PQ-R the transmission of routing information is totally data-driven. The learning and adaptation rate we used were the same as used by the algorithm's authors (Boyan & Littman, 1994; Choi & Yeung, 1996).

|  | AntNet | OSPF & SPF | BF | Q-R & PQ-R |
|---|---|---|---|---|
| Packet size (byte) | $24 + 8N_h$ | $64 + 8|\mathcal{N}_n|$ | $24 + 12N$ | 12 |
| Packet elaboration time (msec) | 3 | 6 | 2 | 3 |

Table 1: Routing packets characteristics for the implemented algorithms (except for the Daemon algorithm, which does not generate routing packets). $N_h$ is the incremental number of hops made by the forward ant, $|\mathcal{N}_n|$ is the number of neighbors of node $n$, and $N$ is the number of network nodes.

---

13. The knowledge about the characteristics of the incoming CBR or VBR bit streams is of fundamental importance in networks able to deliver Quality of Service. It is only on the basis of this knowledge that the network can accept/refuse the session requests, and, in case of acceptance, allocate/reserve necessary resources.

Concerning AntNet, we observed that the algorithm is very robust to internal parameters tuning. We did not finely tune the parameter set, and we used the same set of values for all the different experiments we ran. Most of the settings we used have been previously given in the text at the moment the parameter was discussed and they are not reported in this section. The ant generation interval at each node was set to 0.3 seconds. In Section 7.4 it will be shown the robustness of AntNet with respect to this parameter. Regarding the parameters of the statistical model, the value of $\eta$, weighting the number of the samples considered in the model (Equation 1), has been set to 0.005, the $c$ factor for the expression of $|\mathcal{W}|_{max}$ (sect. 4.2) has been put equal to 0.3, and the confidence level factor $z$ (sect. 4.2) equal to 1.70, meaning a confidence level of approximately 0.95.

## 7. Results

Experiments reported in this section compare AntNet with the competing routing algorithms described in Section 5. We studied the performance of the algorithms for increasing traffic load, examining the evolution of the network status toward a saturation condition, and for temporary saturation conditions.

- Under *low load conditions*, all algorithms tested have similar performance. In this case, also considering the huge variability in the possible traffic patterns, it is very hard to assess whether an algorithm is significantly better than another or not.

- Under *high, near saturation, loads*, all the tested algorithms are able to deliver the offered throughput in a quite similar way, that is, in most of the cases all the generated traffic is routed without big losses. On the contrary, the study of packet delay distributions shows remarkable differences among the different algorithms. To present simulation results regarding packet delays we decided either to report the whole empirical distribution or to use the 90-th percentile statistic, which allows one to compare the algorithms on the basis of the upper value of delay they were able to keep the 90% of the correctly delivered packets. In fact, packet delays can be spread over a wide range of values. This is an intrinsic characteristics of data networks: packet delays can range from very low values for sessions open between adjacent nodes connected by fast links, to much higher values in the case of sessions involving nodes very far apart connected by many slow links. Because of this, very often the empirical distribution of packet delays cannot be meaningfully parametrized in terms of mean and variance, and the 90-th percentile statistic, or still better the whole empirical distribution, are much more meaningful.

- Under *saturation* there are packet losses and/or packet delays that become too big, cause all the network operations to slow down. Therefore, saturation has to be only a temporary situation. If it is not, structural changes to the network characteristics, like adding new and faster connection lines, rather than improvements of the routing algorithm, should be in order. For these reasons, we studied the responsiveness of the algorithms to traffic loads causing only a temporary saturation.

All reported data are averaged over 10 trials lasting 1000 virtual seconds of simulation time. One thousand seconds represents a time interval long enough to expire all transients and to get enough statistical data to evaluate the behavior of the routing algorithm. Before being fed with data traffic, the algorithms are given 500 preliminary simulation seconds with no data traffic to build initial routing tables. In this way, each algorithm builds the routing tables according to its own "vision" about minimum cost paths. Results for throughput are reported as average values without an associated measure of variance. The inter-trial variability is in fact always very low, a few percent of the average value.

Parameter values for traffic characteristics are given in the Figure captions with the following meaning (see also previous section): MSIA is the mean of the sessions inter-arrival time distribution for the Poisson (P) case, MPIA stands for the mean of the packet inter-arrival time distribution. In the CBR case, MPIA indicates the fixed packet production rate. HS is the number of hot-spots nodes and MPIA-HS is the equivalent of MPIA for the hot-spot sessions. In the following, when not otherwise explicitly stated, the shape of the session bit streams is assumed to be of GVBR type.

Results for *throughput* and *packet delays* for all the considered network topologies are described in the three following subsections. Results concerning the *network resources utilization* are reported in Section 7.4.

## 7.1 SimpleNet

Experiments with SimpleNet were designed to study how the different algorithms manage to distribute the load on the different possible paths. In these experiments, all the traffic, of F-CBR type, is directed from node 1 to node 6 (see Figure 5), and the traffic load has been set to a value higher than the capacity of a single link, so that it cannot be routed efficiently on a single path.

Results regarding throughput (Figure 8a) in this case strongly discriminate among the algorithms. The type of the traffic workload and the small number of nodes determined significant differences in throughput. AntNet is the only algorithm able to deliver almost all the generated data traffic: its throughput after a short transient phase approaches very closely the level of that delivered by the Daemon algorithm. PQ-R attains a steady value approximately 15% inferior to that obtained by AntNet. The other algorithms behave very poorly, stabilizing on values of about 30% inferior to those provided by AntNet. In this case, it is rather clear that AntNet is the only algorithm able to exploit at best all the three available paths (1-8-7-6, 1-3-5-6, 1-2-4-5-6) to distribute the data traffic without inducing counterproductive oscillations. The utilization of the routing tables in a probabilistic way also by data packets in this case plays a fundamental role in achieving higher quality results. Results for throughput are confirmed by those for packet delays, reported in the graph of Figure 8b. The differences in the empirical distributions for packet delays reflect approximatively the same proportions as evidenced in the throughput case.

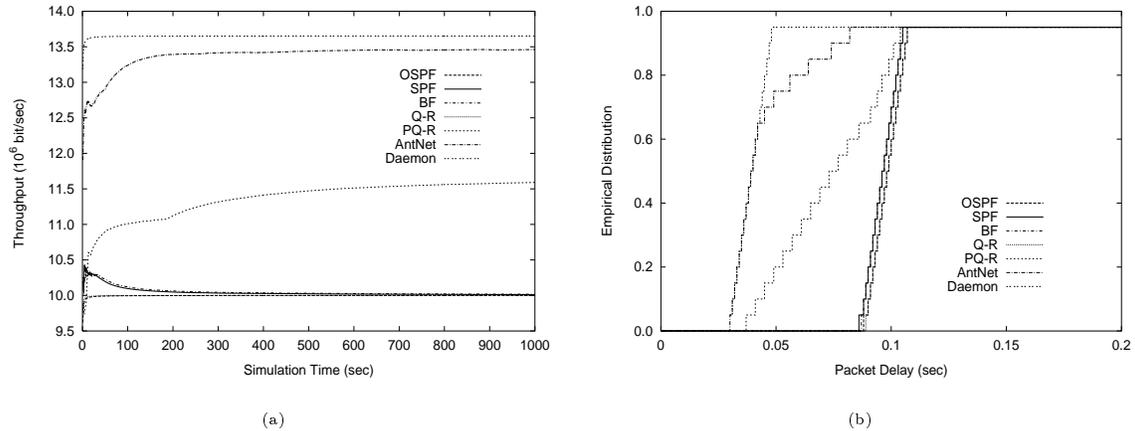

(a)

(b)

Figure 8: *SimpleNet*: Comparison of algorithms for F-CBR traffic directed from node 1 to node 6 (MPIA = 0.0003 sec). (a) Throughput, and (b) packet delays empirical distribution.

## 7.2 NSFNET

We carried out a wide range of experiments on NSFNET using UP, RP, UP-HS and TMPHS-UP traffic patterns. In all the cases considered, differences in throughput are of minor importance with respect to those shown by packet delays. For each one of the UP, RP and UP-HS cases we ran five distinct groups of ten trial experiments, gradually increasing the generated workload (in terms of reducing the session inter-arrival time). As explained above, we studied the behavior of the algorithms when moving the traffic load towards a saturation region.

In the UP case, differences in throughput (Figure 9a) are small: the best performing algorithms are BF and SPF, which can attain performance of only about 10% inferior to those of Daemon and of the same amount better than those of AntNet, Q-R and PQ-R,[14] while OSPF behaves slightly better than these last ones. Concerning delays (Figure 9b) the situation is rather different, as can be seen by the fact that all the algorithms but AntNet have been able to produce a slightly higher throughput at the expenses of much worse results for packet delays. This trend in packet delays was confirmed by all the experiments we ran. OSPF, Q-R and PQ-R show really poor results (delays of order 2 or more seconds are very high values, even if we are considering the 90-th percentile of the distribution), while BF and SPF behave in a similar way with performance of order 50% worse than those obtained by AntNet and of order 65% worse than Daemon.

---

14. It is worth remarking that in these and in some of the experiments presented in the following, PQ-R's performance is slightly worse than that of Q-R. This seems to be in contrast with the results presented by the PQ-R's authors in the article where they introduced PQ-R (Choi & Yeung, 1996). We think that this behavior is due to the fact that (i) their link recovery rate matches a discrete-time system while in our simulator time is a continuous variable, and (ii) the experimental and simulation conditions are rather different (in their article it is not specified the way they produced traffic patterns and they did not implement a realistic network simulator).

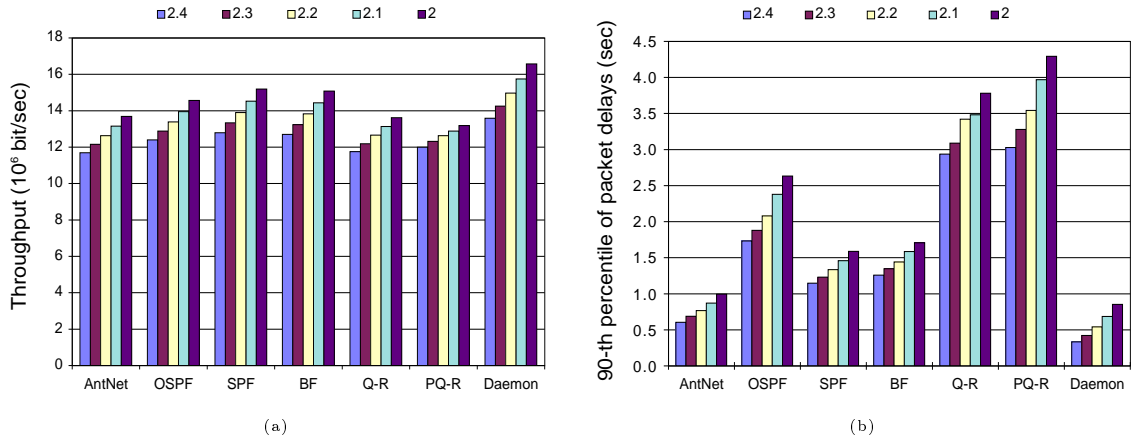

Figure 9: *NSFNET*: Comparison of algorithms for increasing load for UP traffic. The load is increased reducing the MSIA value from 2.4 to 2 seconds (MPIA = 0.005 sec). (a) Throughput, and (b) 90-th percentile of the packet delays empirical distribution.

In the RP case (Figure 10a), throughputs generated by AntNet, SPF and BF are very similar, although AntNet has a slightly better performance. OSPF and PQ-R behave only slightly worse while Q-R is the worst algorithm. Daemon is able to obtain only slightly better results than AntNet. Again, looking at packet delays results (Figure 10b) OSPF, Q-R and PQ-R perform very badly, while SPF shows results a bit better than those of BF but of order 40% worse than those of AntNet. Daemon is in this case far better, which indicates that the testbed was very difficult.

For the case of UP-HS load, throughputs (Figure 11a) for AntNet, SPF, BF, Q-R and Daemon are very similar, while OSPF and PQ-R clearly show much worse results. Again (Figure 11b), packet delays results for OSPF, Q-R and PQ-R are much worse than those of the other algorithms (they are so much worse that they do not fit in the scale chosen to make clear differences among the other algorithms). AntNet is still the best performing algorithm. In this case, differences with SPF are of order 20% and of 40% with respect to BF. Daemon performs about 50% better than AntNet and scales much better than AntNet, which, again, indicates the testbed was rather difficult.

The last graph for NSFNET shows how the algorithms behave in the case of a TMPHS-UP situation (Figure 12). At time $t = 400$ four hot spots are turned on and superimposed to the existing light UP traffic. The transient is kept on for 120 seconds. In this case, only one, typical, situation is reported in detail to show the answer curves. Reported values are the "instantaneous" values for throughput and packet delays computed as the average over 5 seconds moving windows. All algorithms have a similar very good performance as far as throughput is concerned, except for OSPF and PQ-R, which lose a few percent of the packets during the transitory period. The graph of packet delays confirms previous results: SPF and BF have a similar behavior, about 20% worse than AntNet and 45% worse than Daemon. The other three algorithms show a big out-of-scale jump, being not able to properly dump the sudden load increase.

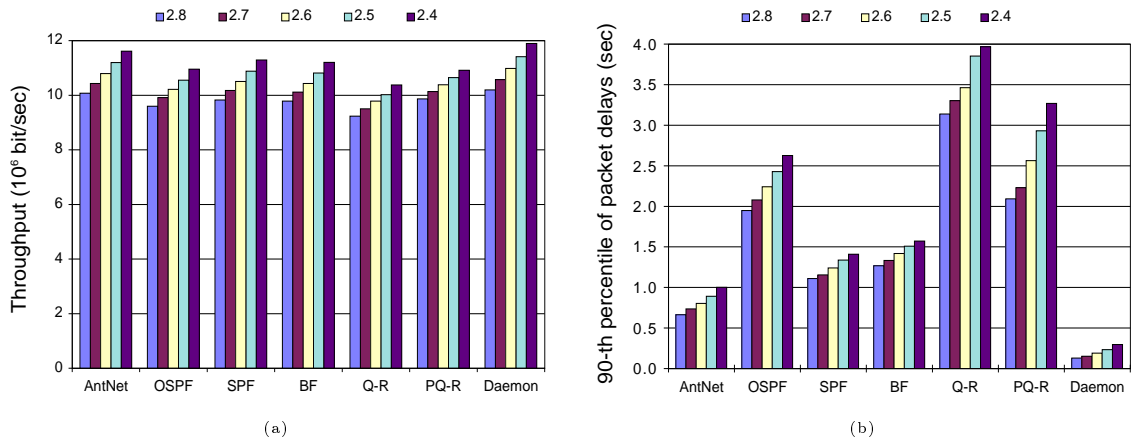

(a)                                                    (b)

Figure 10: *NSFNET*: Comparison of algorithms for increasing load for RP traffic. The load is increased reducing the MSIA value from 2.8 to 2.4 seconds (MPIA = 0.005 sec). (a) Throughput, and (b) 90-th percentile of the packet delays empirical distribution.

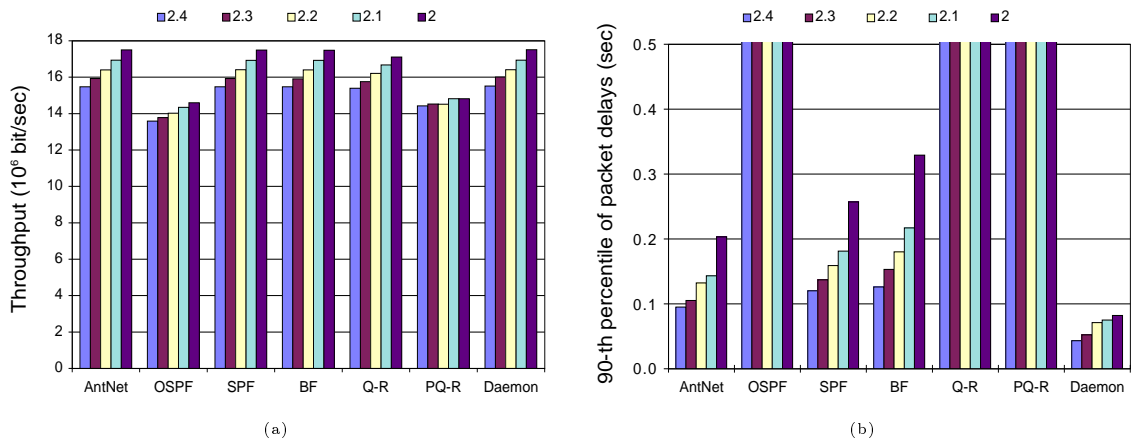

(a)                                                    (b)

Figure 11: *NSFNET*: Comparison of algorithms for increasing load for UP-HS traffic. The load is increased reducing the MSIA value from 2.4 to 2.0 seconds (MPIA = 0.3 sec, HS = 4, MPIA-HS = 0.04 sec). (a) Throughput, and (b) 90-th percentile of the packet delays empirical distribution.

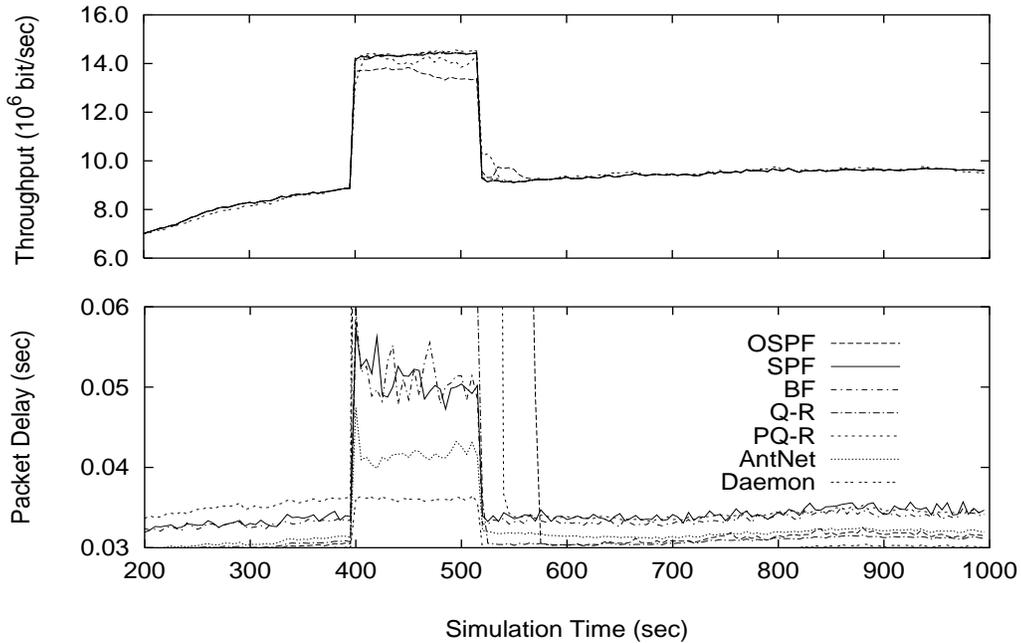

Figure 12: *NSFNET*: Comparison of algorithms for transient saturation conditions with TMPHS-UP traffic (MSIA = 3.0 sec, MPIA = 0.3 sec, HS = 4, MPIA-HS = 0.04). (a) Throughput, and (b) packet delays averaged over 5 seconds moving windows.

## 7.3 NTTnet

The same set of experiments run on the NSFNET have been repeated on NTTnet. In this case the results are even sharper than those obtained with NSFNET: AntNet performance is much better that of all its competitors.

For the UP, RP and UP-HS cases, differences in throughput are not significant (Figures 13a, 14a and 15a). All the algorithms, with the OSPF exception, practically behave in the same way as the Daemon algorithm. Concerning delays (Figures 13b, 14b and 15b), differences between AntNet and each of its competitors are of one order of magnitude. AntNet keeps delays at low values, very close to those obtained by Daemon, while SPF, BF, Q-R and PQ-R perform poorly and OSPF completely collapses.
In the UP and RP cases (Figures 13b and 14b) SPF and BF performs similarly, even if SPF shows slightly better results, and about 50% better than Q-R and PQ-R.

In the UP-HS case, again, SPF and BF show similar results, while Q-R performs comparably but in a much more irregular way and PQ-R can keep delays about 30% lower. OSPF, which is the worse algorithm in this case, shows an interesting behavior. The increase in the generated data throughput determines a decrease or a very slow increase in the delivered throughput while delays decrease (Figure 15a and 15b). In this case the load was too high for the algorithm and the balance between the two, conflicting, objectives, throughput and packet delays, showed an inverse dynamics: having a lot of packet losses made it possible for the surviving packets to obtain lower trip delays.

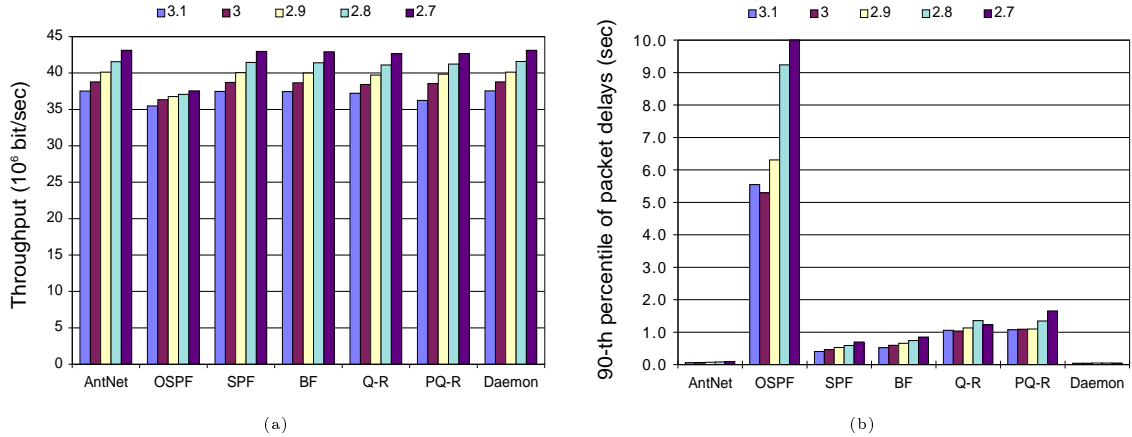

(a)

(b)

Figure 13: *NTTnet*: Comparison of algorithms for increasing load for UP traffic. The load is increased reducing the MSIA value from 3.1 to 2.7 seconds (MPIA = 0.005 sec). (a) Throughput, and (b) 90-th percentile of the packet delays empirical distribution.

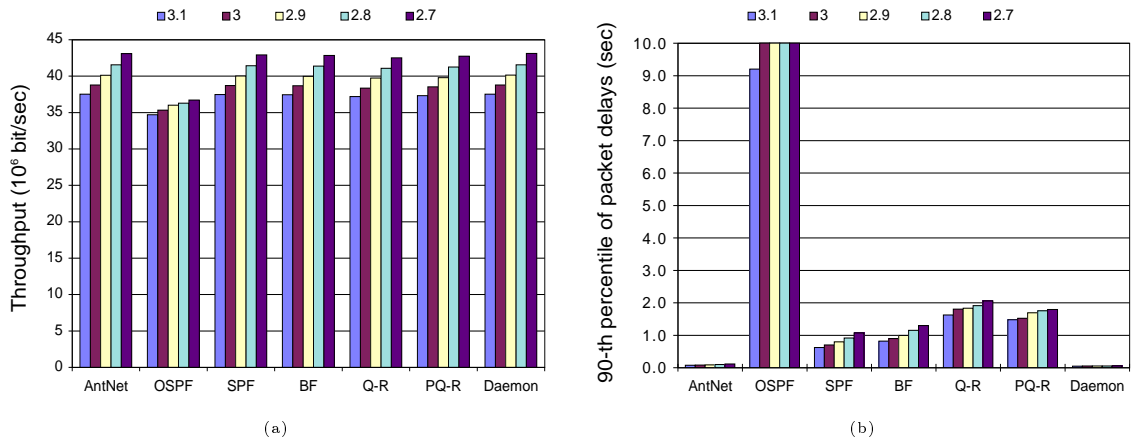

Figure 14: *NTTnet*: Comparison of algorithms for increasing load for RP traffic. The load is increased reducing the MSIA value from 3.1 to 2.7 seconds (MPIA = 0.005 sec). (a) Throughput, and (b) 90-th percentile of the packet delays empirical distribution.

The TMPHS-UP experiment (Figure 16), concerning sudden load variation, confirms the previous results. OSPF is not able to follow properly the variation both for throughput and delays. All the other algorithms are able to follow the sudden increase in the offered throughput, but only AntNet (and Daemon) show a very regular behavior. Differences in packet delays are striking. AntNet performance is very close to those obtained by Daemon (the curves are practically superimposed at the scale used in the Figure). Among the other algorithms, SPF and BF are the best ones, although their response is rather irregular and, in any case, much worse than AntNet's. OSPF and Q-R are out-of-scale and show a very delayed recovering curve. PQ-R, after a huge jump, which takes the graph out-of-scale in the first 40 seconds after hot spots are turned on, shows a trend approaching those of BF and SPF.

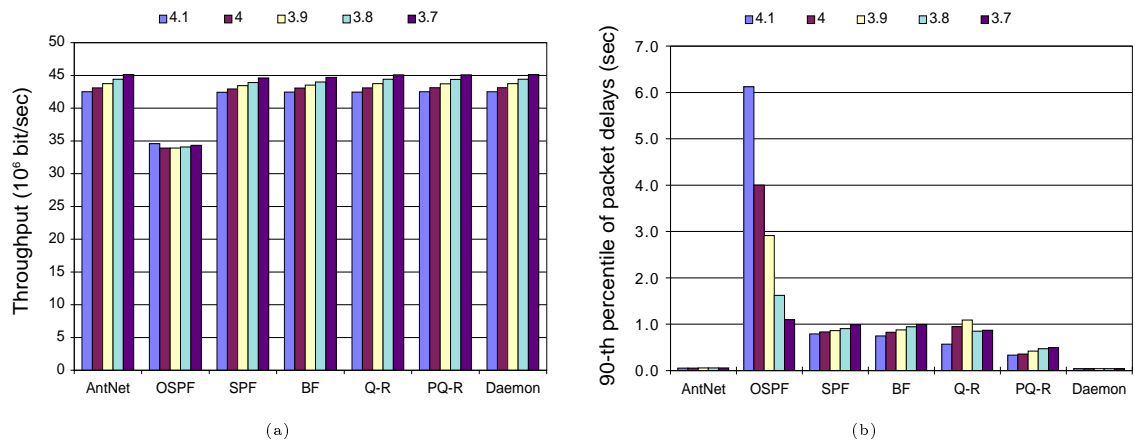

Figure 15: *NTTnet*: Comparison of algorithms for increasing load for UP-HS traffic. The load is increased reducing the MSIA value from 4.1 to 3.7 seconds (MPIA = 0.3 sec, HS = 4, MPIA-HS = 0.05 sec). (a) Throughput, and (b) 90-th percentile of the packet delays empirical distribution.

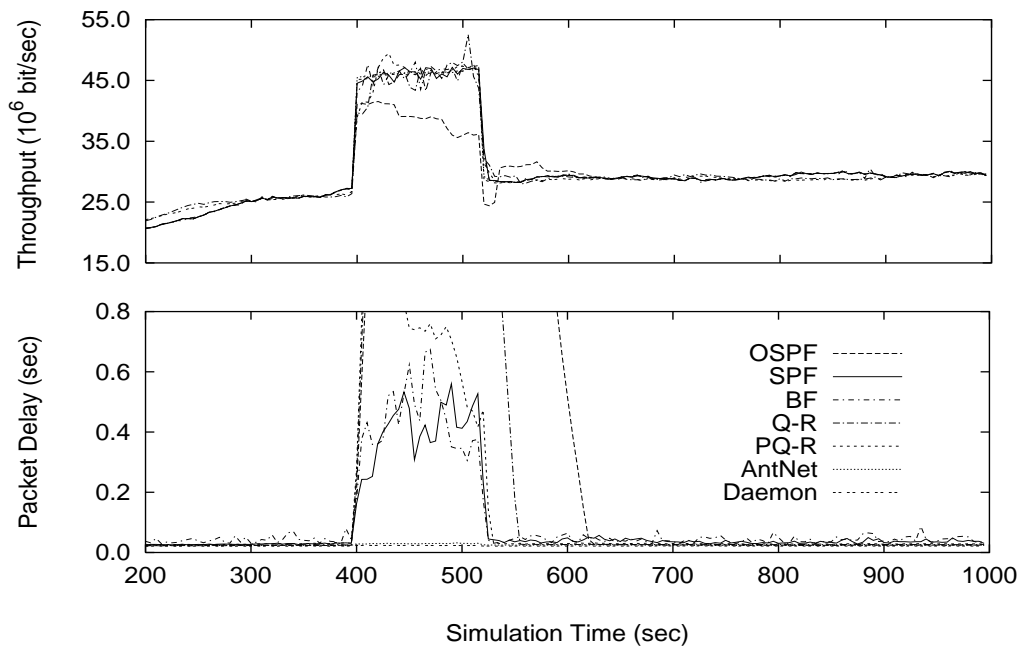

Figure 16: *NTTnet*: Comparison of algorithms for transient saturation conditions with TMPHS-UP traffic (MSIA = 4.0 sec, MPIA = 0.3 sec, HS = 4, MPIA-HS = 0.05). (a) Throughput, and (b) packet delays averaged over 5 seconds moving windows.

## 7.4 Routing Overhead

Table 2 reports results concerning the overhead generated by routing packets. For each algorithm the network load generated by the routing packets is reported as the ratio between the bandwidth occupied by the routing packets and the total available network bandwidth. Each row in the table refers to a previously discussed experiment (Figs. 8 to 11 and 13 to 15). Routing overhead is computed for the experiment with the heaviest load in the increasing load series.

|  | AntNet | OSPF | SPF | BF | Q-R | PQ-R |
|---|---|---|---|---|---|---|
| SimpleNet - F-CBR | 0.33 | 0.01 | 0.10 | 0.07 | 1.49 | 2.01 |
| NSFNET - UP | 2.39 | 0.15 | 0.86 | 1.17 | 6.96 | 9.93 |
| NSFNET - RP | 2.60 | 0.15 | 1.07 | 1.17 | 5.26 | 7.74 |
| NSFNET - UP-HS | 1.63 | 0.15 | 1.14 | 1.17 | 7.66 | 8.46 |
| NTTnet - UP | 2.85 | 0.14 | 3.68 | 1.39 | 3.72 | 6.77 |
| NTTnet - RP | 4.41 | 0.14 | 3.02 | 1.18 | 3.36 | 6.37 |
| NTTnet - UP-HS | 3.81 | 0.14 | 4.56 | 1.39 | 3.09 | 4.81 |

Table 2: *Routing Overhead*: ratio between the bandwidth occupied by the routing packets and the total available network bandwidth. All data are scaled by a factor of $10^{-3}$.

All data are scaled by a factor of $10^{-3}$. The data in the table show that the routing overhead is negligible for all the algorithms with respect to the available bandwidth. Among the adaptive algorithms, BF shows the lowest overhead, closely followed by SPF. AntNet generates a slightly bigger consumption of network resources, but this is widely compensated by the much higher performance it provides. Q-R and PQ-R produce an overhead a bit higher than that of AntNet. The routing load caused by the different algorithms is a function of many factors, specific of each algorithm. Q-R and PQ-R are data-driven algorithms: if the number of data packets and/or the length of the followed paths (because of topology or bad routing) grows, so will the number of generated routing packets. BF, SPF and OSPF have a more predictable behavior: the generated overhead is mainly function of the topological properties of the network and of the generation rate of the routing information packets. AntNet produces a routing overhead depending on the ants generation rate and on the length of the paths they travel.

The ant traffic can be roughly characterized as a collection of additional traffic sources, one for each network node, producing very small packets (and related acknowledgement packets) at constant bit rate with destinations matching the offered data traffic. On average ants will travel over rather "short" paths and their size will grow of only 8 bytes at each hop. Therefore, each "ant routing traffic source" represents a very light additional traffic source with respect to network resources when the ant launching rate is not excessively high. In Figure 17, the sensitivity of AntNet with respect to the ant launching rate is reported.

For a sample case of a UP data traffic model on NSFNET (previously studied in Figure 9) the interval $\Delta g$ between two consecutive ant generations is progressively decreased ($\Delta g$ is the same for all nodes). $\Delta g$ values are sampled at constant intervals over a logarithmic scale ranging from about 0.006 to 25 seconds. The lower, dashed, curve interpolates the generated routing overhead expressed, as before, as the fraction of the available network bandwidth used by routing packets. The upper, solid, curve plots the data for the obtained power normalized to its highest value, where the power is defined as the ratio between the delivered throughput and the packet delay. The value used for delivered throughput is the throughput value at time 1000 averaged over ten trials, while for packet delay we used the 90-th percentile of the empirical distribution.

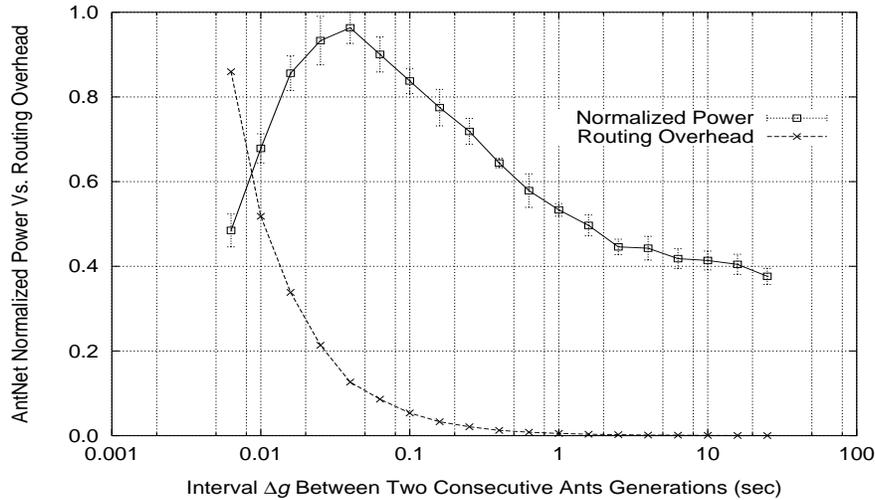

Figure 17: *AntNet* normalized power *vs.* routing overhead. Power is defined as the ratio between delivered throughput and packet delay.

In the figure, we can see how an excessively small $\Delta g$ causes an excessive growth of the routing overhead, with consequent reduction of the algorithm power. Similarly, when $\Delta g$ is too big, the power slowly diminishes and tends toward a plateau because the number of ants is not enough to generate and maintain up-to-date statistics of the network status. In the middle of these two extreme regions a wide range of $\Delta g$ intervals gives raise to similar, very good power values, while, at the same time, the routing overhead quickly falls down toward negligible values. This figure strongly confirms our previous assertion about the robustness of AntNet's internal parameter settings.

## 8. Discussion

In AntNet, the continual on-line construction of the routing tables is the emergent result of a collective learning process. In fact, each forward-backward agent pair is complex enough to find a good route and to adapt the routing tables for a single source-destination path, but it cannot solve the global routing optimization problem. It is the interaction between the agents that determines the emergence of a global effective behavior from the network performance point of view. Ants cooperate in their problem-solving activity by communicating in an indirect and non-coordinated way. Each agent acts independently. Good routes are discovered by applying a policy that is a function of the information accessed through the network nodes visited, and the information collected about the route is eventually released on the same nodes. Therefore, the inter-agent communication is mediated in an explicit and implicit way by the "environment", that is, by the node's data structures and by the traffic patterns recursively generated by the data packets' utilization of the routing tables. This communication paradigm, called stigmergy, matches well the intrinsically distributed nature of the routing problem. Cooperation among agents goes on at two levels: (a) by modifications of the routing tables, and (b) by modifications of local models that determine the way the ants' performance is evaluated. Modifications of the routing tables directly affect the routing decisions of following ants towards the same destination, as well as the routing of data, which, in turn, influences the rate of arrival of other ants towards any destination. It is interesting to remark that the used stigmergy paradigm makes the AntNet's mobile agents very flexible from a software engineering point of view. In this perspective, once the interface with the node's data structure is defined, the internal policy of the agents can be transparently updated. Also, the agents could be exploited to carry out multiple concurrent tasks (e.g., collecting information for distributed network management using an SNMP-like protocol or for Web data-mining tasks).

As shown in the previous section, the results we obtained with the above stigmergetic model of computation are excellent. In terms of throughput and average delay, AntNet performs better than both classical and recently proposed routing algorithms on a wide range of experimental conditions. Although this is very interesting *per se*, in the following we try to justify AntNet superior performance by highlighting some of its characteristics and by comparing them with those of the competing algorithms. We focus on the following main aspects:

- AntNet can be seen as a particular instance of a parallel Monte Carlo simulation system with biased exploration. All the other algorithms either do not explore the net or their exploration is local and tightly connected to the flux of data packets.

- The information AntNet maintains at each node is more complete and organized in a less critical way than that managed by the other algorithms.

- AntNet does not propagate local estimates to other nodes, while all its competitors do. This mechanism makes the algorithm more robust to locally wrong estimates.

- AntNet uses probabilistic routing tables, which have the triple positive effect of better redistributing data traffic on alternative routes, of providing ants with a built-in exploration mechanism and of allowing the exploitation of the ants' arrival rate to assign cumulative reinforcements.

- It was experimentally observed that AntNet is much more robust than its competitors to the frequency with which routing tables are updated.

- The structure of AntNet allows one to draw some parallels with some well-known reinforcement learning (RL) algorithms. The characteristics of the routing problem, that can be seen as a distributed time-varying RL problem (see sect. 2.2), determines a departure of AntNet from the structure of classical RL algorithms.

These aspects of AntNet are discussed in more detail in the following.

## 8.1 AntNet as an on-line Monte Carlo system with biased exploration

The AntNet routing system can be seen as a collection of mobile agents collecting data about the network status by concurrently performing on-line Monte Carlo simulations (Rubistein, 1981; Streltsov & Vakili, 1996). In Monte Carlo methods, repeated experiments with stochastic transition components are run to collect data about the statistics of interest. Similarly, in AntNet ants explore the network by performing random experiments (i.e., building paths from source to destination nodes using a stochastic policy dependent on the past and current network states), and collect on-line information on the network status. A built-in variance reduction effect is determined (i) by the way ants' destinations are assigned, biased by the most frequently observed data's destinations, and (ii) by the way the ants' policy makes use of current and past traffic information (that is, inspection of the local queues' status and probabilistic routing tables). In this way, the explored paths match the most interesting paths from a data traffic point of view, which results in a very efficient variance reduction effect in the stochastic sampling of the paths. Differently from usual off-line Monte Carlo systems, in AntNet the state space sampling is performed on-line, that is, the sampling of the statistics and the controlling of the non-stationary traffic process are performed concurrently.

This way of exploring the network concurrently with data traffic is very different from what happens in the other algorithms where, either there is no exploration at all (OSPF, SPF and BF), or exploration is both tightly coupled to data traffic and of a local nature (Q-R and PQ-R). Conveniently, as was shown in Section 7.4, the extra traffic generated by exploring ants is negligible for a wide range of values, allowing very good performance.

## 8.2 Information management at each network node

Key characteristics of routing algorithms are the type of information used to build/update routing tables and the way this information is propagated. All the algorithms (except the static OSPF) make use at each node of two main components: a local model $\mathcal{M}$ of some cost measures and a routing table $\mathcal{T}$. SPF and BF use $\mathcal{M}$ to estimate smoothed averages of the local link costs, that is, of the distances to the neighbor nodes. In this case, $\mathcal{M}$ is a local model maintaining estimates of only local components. In Q-R the local model is fictitious because the raw transition time is directly used as a value to update $\mathcal{T}$. PQ-R uses a slightly more sophisticated model with respect to Q-R, storing also a measure of the link utilization. All these algorithms propagate part of their local information to the other nodes, which, in turn, make use of it to update their routing tables and to build a global view of the network. In SPF and BF the content of each $\mathcal{T}$ is updated, at regular intervals, by a "memoryless strategy": the new entries do not depend on the old values, that are discarded. Therefore, the whole adaptive component of the routing system is represented by the model $\mathcal{M}$. Otherwise, in Q-R and PQ-R the adaptive content of $\mathcal{M}$ is almost negligible and the adaptive component of the algorithm is represented by the smoothed average carried out by the Q-learning-like rule. AntNet shows characteristics rather different from its competitors: its model $\mathcal{M}$ contains a memory-based local perspective of the global status of the network. The content of $\mathcal{M}$ allow the reinforcements to be weighted on the basis of a rich statistical description of the network dynamics as seen by the local node. These reinforcements are used to update the routing table, the other adaptive component maintained at the node. The $\mathcal{T}$ updates are carried out in an asynchronous way and as a function of their previous values. Moreover, while $\mathcal{T}$ is used in a straightforward probabilistic way by the data packets, traveling ants select the next node by using both $\mathcal{T}$, that is, an adaptive representation of the past policy, and a model of the current local link queues, that is, an instantaneous representation of the node status. It is evident that AntNet builds and uses more information than its competitors: two different memory-based components and an instantaneous predictor are used and combined at different levels. Moreover, in this way AntNet robustly redistributes among these completely local components the criticality of all the estimates and decisions.

## 8.3 AntNet's robustness to wrong estimates

As remarked above, AntNet, differently from its competitors, does not propagate local estimates to other nodes. Each node routing table is updated independently, by using local information and the ants' experienced trip time. Moreover, (i) each ant experiment affects only one entry in the routing table of the visited nodes, the one relative to the ant's destination, and, (ii) the local information is built from the "global" information collected by traveling ants, implicitly reducing in this way the variance in the estimates. These characteristics make AntNet particularly robust to wrong estimates. On the contrary, in all the other algorithms a locally wrong estimate will be propagated to all other nodes and will be used to compute estimates to many different destinations. How bad this is for the algorithm performance depends on how long the wrong estimate effect lasts. In particular, this will be a function of the time window over which estimates are computed for SPF and BF, and of the learning parameters for Q-R and PQ-R.

## 8.4 AntNet's probabilistic use of routing tables to route data packets

All the tested algorithms but AntNet use deterministic routing tables.[15] In these algorithms, entries in the routing tables contain distance/time estimates to the destinations. These estimates can provide misleading information if the algorithm is not fast enough to follow the traffic fluctuations, as can be the case under heavy load conditions. Instead, AntNet routing tables have probabilistic entries that, although reflecting the goodness of a particular path choice with respect to the others available, do not force the data packets to choose the perceived best path. This has the positive effect of allowing a better balancing of the traffic load on different paths, with a resulting better utilization of the resources (as was shown in particular in the experiments with the SimpleNet). As remarked at the end of Section 4.1, the intrinsic probabilistic structure of the routing tables and the way they are updated allow AntNet to exploit the ant's arrival rate as a way to assign implicit (cumulative) reinforcements to discovered paths. It is not obvious how the same effect could be obtained by using routing tables containing distance/time estimates and using this estimates in a probabilistic way. In fact, in this case each new trip time sample would modify the statistical estimate that would simply oscillate around its expected value without inducing an arrival-dependent cumulative effect.

---

15. Singh, Jaakkola, and Jordan (1994) showed that stochastic policies can yield higher performance than deterministic policies in the case of an incomplete access to the state information of the environment. In (Jaakkola, Singh, & Jordan, 1995), the same authors developed a Monte-Carlo-based stochastic policy evaluation algorithm, confirming the usefulness of the Monte-Carlo approach, used in AntNet too, to deal with incomplete information problems.

Probabilistic routing tables provide some remarkable additional benefits: (a) they give to the ants a built-in exploration method in discovering new, possibly better, paths, and (b) since ants and data routing are independent in AntNet, the exploration of new routes can continue while, at the same time, data packets can exploit previously learned, reliable information. It is interesting to note that the use of probabilistic routing tables whose entries are learned in an adaptive way by changing on positive feedback and ignoring negative feedback, is reminiscent of older automata approaches to routing in telecommunications networks. In these approaches, a learning automaton is usually placed on each network node. An automaton is defined by a set of possible actions and a vector of associated probabilities, a continuous set of inputs and a learning algorithm to learn input-output associations. Automata are connected in a feedback configuration with the environment (the whole network), and a set of penalty signals from the environment to the actions is defined. Routing choices and modifications to the learning strategy are carried out in a probabilistic way and according to the network conditions (see for example (Nedzelnitsky & Narendra, 1987; Narendra & Thathachar, 1980)). The main difference lies in the fact that in AntNet the ants are part of the environment itself, and they actively direct the learning process towards the most interesting regions of the search space. That is, the whole environment plays a key, active role in learning good state-action pairs.

## 8.5 AntNet robustness to routing table update frequency

In BF and SPF the broadcast frequency of routing information plays a critical role, particularly so for BF, which has only a local representation of the network status. This frequency is unfortunately problem dependent, and there is no easy way to make it adaptive, while, at the same time, avoiding large oscillations. In Q-R and PQ-R, routing tables updating is data driven: only those Q-values belonging to pairs $(i, j)$ of neighbor nodes visited by packets are updated. Although this is a reasonable strategy given that the exploration of new routes could cause undesired delays to data packets, it causes delays in discovering new good routes, and is a great handicap in a domain where good routes could change all the time. In OSPF, in which routing tables are not updated, we set static link costs on the basis of their physical characteristics. This lack of an adaptive metric is the main reason of the poor performance of OSPF (as remarked in Section 5, we slightly penalized OSPF with respect to its real implementations, where additional heuristic knowledge about traffic patterns is used by network administrators to set link costs). In AntNet, we experimentally observed the robustness to changes in the ants' generation rate: for a wide range of generation rates, rather independent of the network size, the algorithm performance is very good and the routing overhead is negligible (see Section 7.4).

## 8.6 AntNet and reinforcement learning

The characteristics of the routing problem allow one to interpret it as a distributed, stochastic time-varying RL problem. This fact, as well as the structure of AntNet, make it natural to draw some parallels between AntNet and classical RL approaches. It is worth remarking that those RL problems that have been most studied, and for which algorithms have been de-veloped, are problems where, unlike routing, assumptions like Markovianity or stationarity of the process considered are satisfied. The characteristics of the adaptive routing problem make it very difficult and not well suited to be solved with usual RL algorithms. This fact, as we explain below, determines a departure of AntNet from classical RL algorithms.

A first way to relate the structure of AntNet to that of a (general) RL algorithm is connected to the way the outcomes of the experiments, the trip times $T_{k \to d}$, are processed. The transformation from the raw values $T_{k \to d}$ to the more refined reinforcements $r$ are reminiscent of what happens in Actor-Critic systems (Barto, Sutton, & Anderson, 1983): the raw reinforcement signal is processed by a critic module, which is learning a model (the node's component $\mathcal{M}$) of the underlying process, and then is fed to the learning system (the routing table $\mathcal{T}$) transformed into an evaluation of the policy followed by the ants. In our case, the critic is both adaptive, to take into account the variability of the traffic process, and rather simple, to meet computational requirements.

Another way of seeing AntNet as a classical RL system is related to its interpretation as a parallel replicated Monte Carlo (MC) system. As was shown by Singh and Sutton (1996), a first-visit MC (only the first visit to a state is used to estimate its value during a trial) simulation system is equivalent to a batch temporal difference (TD) method with replacing traces and decay parameter $\lambda=1$. Although AntNet is a first-visit MC simulation system, there are some important differences with the type of MC used by Singh and Sutton (and in other RL works), mainly due to the differences in the considered class of problems. In AntNet, outcomes of experiments are both used to update local models able to capture the variability of the whole network status (only partially observable) and to generate a sequence of stochastic policies. On the contrary, in the MC system considered by Singh and Sutton, outcomes of the experiments are used to compute (reduced) maximum-likelihood estimates of the expected mean and variance of the states' returns (i.e., the total reward following a visit of a state) of a Markov chain. In spite of these differences, the weak parallel with TD($\lambda$) methods is rather interesting, and allows to highlight an important difference between AntNet and its competitors (and general TD methods): in AntNet, following the generation of a stochastic transition chain by the forward ant, there is no back-chaining of the information from one state (i.e., a triple {current node, destination node, next hop node}) to its predecessors. Each state is rewarded only on the basis of the ant's trip time information strictly relevant to it. This approach is completely different from that followed by (TD methods) Q-R, PQ-R, BF and, in a different perspective, by SPF. In fact, these algorithms build the distance estimates at each node by using the predictions made at other nodes. In particular, Q-R and PQ-R, which propagate the estimation information only one step back, are precisely distributed versions of the TD(0) class of algorithms. They could be transformed into generic TD($\lambda$), $0 < \lambda \leq 1$, by transmitting backward to all the previously visited nodes the information collected by the routing packet generated after each data hop. Of course, this would greatly increase the routing traffic generated, because it has to be done after each hop of each data packet, making the approach at least very costly, if feasible at all.

In general, using temporal differences methods in the context of routing presents an important problem: the key condition of the method, the self-consistency between the estimates of successive states[16] may not be strictly satisfied in the general case. This is due to the

---

16. For instance, the prediction made at node $k$ about the time to-go to the destination node $d$ should be

fact that (i) the dynamics at each node are related in a highly non-linear way to the dynamics of all its neighbors, (ii) the traffic process evolves concurrently over all the nodes, and (iii) there is a recursive interaction between the traffic patterns and the control actions (that is, the modifications of the routing tables). This aspect can explain in part the poor performance of the pure TD(0) algorithms Q-R and PQ-R.

## 9. Related Work

Algorithms based on the ant colony metaphor were inspired by the ant colony foraging behavior (Beckers et al., 1992). These were first proposed by Dorigo (1992), Colorni et al. (1991) and Dorigo et al. (1991, 1996) and were applied to the traveling salesman problem (TSP). Apart from the natural metaphor, the idea behind that first application was similar to the one presented in this paper: a set of agents that repeatedly run Monte Carlo experiments whose outcomes are used to change the estimates of some variables used by subsequent ants to build solutions. In *ant-cycle*, one of the first ant-based algorithms, a value called "pheromone trail" is associated to each edge of the graph representing the TSP. Each ant builds a tour by exploiting the pheromone trail information as follows. When in node $i$ an ant chooses the next node $j$ to move to among those not visited yet with a probability $P_{ij}$ that is a function of the amount of pheromone trail on the edge connecting $i$ to $j$ (as well as of a local heuristic function; the interested reader can find a detailed description of ant-cycle elsewhere (Dorigo, 1992; Dorigo et al., 1996)). The value of the pheromone trails is updated once all ants have built their tours. Each ant adds to all visited edges a quantity of pheromone trail proportional to the quality of the tour generated (the shorter the tour, the higher the quantity of pheromone trail added). This has an effect very similar to AntNet's increase of routing tables probabilities, since a higher pheromone trail on a particular edge will increase its probability of being chosen in the future. There are obviously many differences between ant-cycle and AntNet, mostly due to the very different types of problems to which they have been applied, a combinatorial optimization problem versus a distributed, stochastic, time varying, real-time problem.

Though the majority of previous applications of ant colony inspired algorithms concern combinatorial optimization problems, there have been recent applications to routing. Schoonderwoerd et al. (1996, 1997) were the first to consider routing as a possible application domain for ant colony algorithms. Their ant-based control (ABC) approach, which is applied to routing in telephone networks, differs from AntNet in many respects. The main differences are a direct consequence of the different network model they considered, which has the following characteristics (see Figure 18): (i) connection links potentially carry an infinite number of full-duplex, fixed bandwidth channels, and (ii) transmission nodes are crossbar switches with limited connectivity (that is, there is no necessity for queue management in the nodes). In such a model, bottlenecks are put on the nodes, and the congestion degree of a network can be expressed in terms of connections still available at each switch. As a result, the network is cost-symmetric: the congestion status over available paths is completely bi-directional. The path $n_0, n_1, n_2, \ldots, n_k$ connecting $n_0$ and $n_k$ will exhibit the

---

additively related to the prediction for the same destination from each one of $k$'s neighbors, being each neighbor one of the ways to go to $d$.

same level of congestion in both directions because the congestion depends only on the state of the nodes in the path. Moreover, dealing with telephone networks, each call occupies exactly one physical channel across the path. Therefore, "calls" are not multiplexed over the links, but they can be accepted or refused, depending on the possibility of reserving a physical circuit connecting the caller and the receiver. All of these modeling assumptions make the problem of Schoonderwoerd et al. very different from the cost-asymmetric routing problem for data networks we presented in this paper. This difference is reflected in many algorithmic differences between ABC and AntNet, the most important of which is that in ABC ants update pheromone trails after each step, without waiting for the completion of an experiment as done in AntNet. This choice, which is reminiscent of the pheromone trail updating strategy implemented in ant-density, another of the first ant colony based algorithms (Dorigo et al., 1991; Dorigo, 1992; Colorni et al., 1991), makes ABC behavior closer to real ants', and was made possible by the cost-symmetry assumption made by the authors.

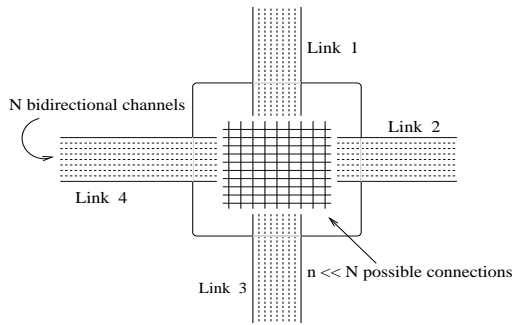

Figure 18: Network node in the telecommunications network model of Schoonderwoerd et al. (1996).

Other differences are that ABC does not use local models to score the ants trip times, nor local heuristic information and ant-private memory to improve the ants decision policies. Also, it does not recover from cycles and does not use the information contained in all the ant sub-paths.

Because of the different network model used and of the many implementation details tightly bound to the network model, it was impossible for us to re-implement and compare the ABC algorithm with AntNet.

Subramanian, Druschel, and Chen (1997) have proposed an ant-based algorithm for packet-switched nets. Their algorithm is a straightforward extension of Schoonderwoerd et al. system by adding so-called *uniform ants*, an additional exploration mechanism that should avoid a rapid sub-optimal convergence of the algorithm. A limitation of Subramanian et al. work is that, although the algorithm they propose is based on the same cost-symmetry hypothesis as ABC, they apply it to packet-switched networks where this requirement is very often not met.

## 10. Conclusions and Future Work

In this paper, we have introduced AntNet, a novel distributed approach to routing in packet-switched communications networks. We compared AntNet with 6 state-of-the-art routing algorithms on a variety of realistic testbeds. AntNet showed superior performance and robustness to internal parameter settings for almost all the experiments. AntNet's most innovative aspect is the use of stigmergetic communication to coordinate the actions of a set of agents that cooperate to build adaptive routing tables. Although this is not the first application of stigmergy-related concepts to optimization problems (e.g., Dorigo et al., 1991; Dorigo, 1992; Dorigo et al., 1996; Bonabeau, Dorigo, & Théraulaz, 1999), the application presented here is unique in many respects. First, in AntNet, stigmergy-based control is coupled to a model-building activity: information collected by ants is used not only to modify routing tables, but also to build local models of the network status to be used to better direct the routing table modifications. Second, this is the first attempt to evaluate stigmergy-based control on a realistic simulator of communications networks: the used simulator retains many of the basic components of a real routing system. An interesting step forward, in the direction of testing the applicability of the idea presented to real networks, would be to rerun the experiments presented here using a complete Internet simulator. Third, this is also the first attempt to evaluate stigmergy-based control by comparing a stigmergetic algorithm to state-of-the-art algorithms on a realistic set of benchmark problems. It is very promising that AntNet turned out to be the best performing in all the tested conditions.

There are obviously a number of directions in which the current work could be extended, which are listed below.

1) A first, natural, extension of the current work would consider the inclusion in the simulator of flow and congestion control components (with re-transmissions and error management). This inclusion will require a paired tuning of the routing and flow-congestion components, to select the best matching between their dynamics.

2) In AntNet, each forward ant makes a random experiment: it builds a path from a source node $s$ to a destination node $d$. The path is built exploiting the information contained in the probabilistic routing tables and the status of the queues of the visited nodes. While building the path, the ant collects information on the status of the network. This is done by sharing link queues with data packets, and by measuring waiting times of queues and traversal times that will be used as raw reinforcements by backward ants. Since forward ants share queues with data packets, the time required to run an experiment depends on the network load, and is approximately the same as the time $T_{s \to d}$ required for a packet to go from the same source node $s$ to the same destination node $d$. This delays the moment the information collected by forward ants can be distributed by backward ants, and makes it less up-to-date than it could be. A possible improvement in this schema would be to add a model of link-queue depletion to nodes, and to let forward ants use high priority queues to reach their destinations without storing crossing times (for a first step in this direction see Di Caro & Dorigo, 1998). Backward ants would then make the same path, in the opposite direction, as forward ants, but use the queue local models they find on their way to estimate local "virtual" queueing and crossing times. Raw reinforcements, used to update the routing tables, are then computed using these estimates. Clearly, here there is a trade-off between delayed but real information and more recent but estimated information. It will be interesting to see which scheme works better, although we are confident that the local queue models should allow the backward ants to build estimates accurate enough to make the improved system more effective than the current AntNet, at a cost of a little increase in computational complexity at the nodes.

3) As we discussed in Section 8, AntNet is missing one of the main components of classical RL/TD algorithms: there is no back-chaining of information from a state to previous ones, each node policy is learned by using a complete local perspective. An obvious extension of our work would therefore be to study versions of AntNet closer to TD($\lambda$) algorithms.

In this case each node should maintain Q-values expressing the estimate of the distance to each destination via each neighbor. These estimates should be updated by using both the ant trip time outcome and the estimates coming from successive nodes (closer to the destination node) that could be also carried by the backward ant.

4) In this paper we applied AntNet to routing in datagram communications networks. It is reasonable to think that AntNet could be easily adapted to be used for the generation of real-time car route guidance in Dynamic Traffic Assignment (DTA) systems (see for example Yang, 1997). DTA systems exploit currently available and emerging computer, communication, and vehicle sensing technologies to monitor, manage and control the transportation system (the attention is now focused mainly on highway systems) and to provide various levels of information and advice to system users so that they can make timely and informed travel decisions. Therefore, adaptive routing of vehicle traffic presents very similar features to the routing of data packets in communications networks. Moreover, vehicle traffic control systems have the interesting property of a very simplified "transport" layer. In fact, many activities that interfere with routing and that are implemented in the transport layer of communications networks do not exist, or exist only to a limited extent, in vehicles traffic control algorithms. For example, typical transport layer activities like data acknowledgement and retransmission cannot be implemented with real vehicles. Other activities, like flow control, have strong constraints (e.g., people would not be happy to be forbidden to leave their offices for, say, one hour on the grounds that there are already too many cars on the streets!). This makes AntNet still more interesting since it can express its full potential as a routing algorithm.

5) In AntNet, whenever an ant uses a link its desirability (probability) is incremented. Although this strategy, which finds its roots in the ant colony biological metaphor that inspired our work, allowed us to obtain excellent results, it would be interesting to investigate the use of negative reinforcements, even if it can potentially lead to stability problems, as observed by people working on older automata systems. As discussed before, AntNet differs from automata systems because of the active role played by the ants. Therefore, the use of negative reinforcements could show itself to be effective, for example, in reducing the probability of choosing a given link if the ant that used it performed very badly.

## Acknowledgements

This work was supported by a Madame Curie Fellowship awarded to Gianni Di Caro (CEC-TMR Contract N. ERBFMBICT 961153). Marco Dorigo is a Research Associate with the FNRS. We gratefully acknowledge the help received from Tony Bagnall, Nick Bradshaw and George Smith, who proofread and commented an earlier draft of this paper, as well as the many useful comments provided by the three anonymous referees and by Craig Boutilier, the associate editor who managed the review process.

## Appendix A. Optimal and Shortest Path Routing

In this appendix, the characteristics of the two most used routing paradigms, optimal and shortest path routing (introduced in Section 2.1) are summarized:

## A.1 Optimal routing

Optimal routing (Bertsekas & Gallager, 1992) has a network-wide perspective and its objective is to optimize a function of all individual link flows.

Optimal routing models are also called *flow models* because they try to optimize the total mean flow on the network. They can be characterized as multicommodity flow problems, where the commodities are the traffic flows between the sources and the destinations, and the cost to be optimized is a function of the flows, subject to the constraints of flow conservation at each node and positive flow on every link. It is worth observing that the flow conservation constraint can be explicitly stated only if the traffic arrival rate is known.

The routing policy consists of splitting any source-target traffic pair at strategic points, then shifting traffic gradually among alternative routes. This often results in the use of multiple paths for a same traffic flow between an origin-destination pair.

Implicit in optimal routing is the assumption that the main statistical characteristics of the traffic are known and not time-varying. Therefore, optimal routing can be used for static and centralized/decentralized routing. It is evident that this kind of solution suffers all the problems of static routers.

## A.2 Shortest path routing

Shortest path routing (Wang & Crowcroft, 1992) has a source-destination pair perspective. As opposed to optimal routing, there is no global cost function to be optimized. Instead, the route between each node pair is considered by itself and no *a priori* knowledge about the traffic process is required (although of course such knowledge could be fruitfully used). If costs are assigned in a *dynamic* way, based on statistical measures of the link congestion state, a strong feedback effect is introduced between the routing policies and the traffic patterns. This can lead to undesirable oscillations, as has been theoretically predicted and observed in practice (Bertsekas & Gallager, 1992; Wang & Crowcroft, 1992). Some very popular cost metrics take into account queuing and transmission delays, link usage, link capacity and various combination of these measures. The way costs are updated usually involves attempting to reduce big variations considering both long-term and short-term statistics of link congestion states (Khanna & Zinky, 1989; Shankar, Alaettinoğlu, Dussa-Zieger, & Matta, 1992b).

On the other hand, if the costs are *static*, they will reflect both some measure of the expected/wished traffic load over the links and their transmission capacity. Of course, serious loss of efficiency could arise in case of non-stationary conditions or when the a priori assumptions about the traffic patterns are strongly violated in practice.

Considering the different content stored in each routing table, shortest path algorithms can be further subdivided in two classes called *distance-vector* and *link-state* (Steenstrup, 1995; Shankar et al., 1992b). The common behavior of most shortest path algorithms can be depicted as follows.

1. Each node assigns a cost to each of its outgoing links. This cost can be static or dynamic. In the latter case, it is updated in presence of a link failure or on the basis of some observed link-traffic statistics averaged over a defined time-window.

2. Periodically and without a required inter-node synchronization, each node sends to all of its neighbors a packet of information describing its current estimates about some quantities (link costs, distance from all the other nodes, etc.).

3. Each node, upon receiving the information packet, updates its local routing table and executes some class-specific actions.

4. Routing decisions can be made in a deterministic way, choosing the best path indicated by the information stored in the routing table, or adopting a more flexible strategy which uses all the information stored in the table to choose some randomized or alternative path.

In the following, the main features specific to each class are described.

### A.2.1 Distance-vector

Distance-vector algorithms make use of routing tables consisting of a set of triples of the form *(Destination, Estimated Distance, Next Hop)*, defined for all the destinations in the network and for all the neighbor nodes of the considered switch.[17] In this case, the required topological information is represented by the list of the reachable nodes identifiers. The average per node memory occupation is of order $O(Nn)$, where $N$ is the number of nodes in the network and $n$ is the average connectivity degree (i.e., the average number of neighbor nodes considered over all the nodes).

The algorithm works in an iterative, asynchronous and distributed way. The information that every node sends to its neighbors is the list of its last estimates of the distances from itself to all the other nodes in the network. After receiving this information from a neighbor node $j$, the receiving node $i$ updates its table of distance estimates overwriting the entry corresponding to node $j$ with the received values.

Routing decisions at node $i$ are made choosing as next hop node the one satisfying the relationship:

$$arg \min_{j \in \mathcal{N}_i} \{d_{ij} + D_j\}$$

where $d_{ij}$ is the assigned cost to the link connecting node $i$ with its neighbor $j$ and $D_j$ is the estimated shortest distance from node $j$ to the destination.

It can be shown that this process converges in finite time to the shortest paths with respect to the used metric if no link cost changes after a given time (Bertsekas & Gallager, 1992).

The above briefly described algorithm is known in literature as distributed *Bellman-Ford* (Bellman, 1958; Ford & Fulkerson, 1962; Bertsekas & Gallager, 1992) and it is based on the principles of dynamic programming (Bellman, 1957; Bertsekas, 1995). It is the prototype and the ancestor of a wider class of distance-vector algorithms (Malkin & Steenstrup, 1995) developed with the aim of reducing the risk of circular loops and of accelerating the convergence in case of rapid changes in link costs.

---

17. In some cases, only the best estimates are kept at nodes. Therefore, the above triples are defined for all the destinations only.

### A.2.2 Link-state

Link-state algorithms make use of routing tables containing much more information than that used in vector-distance algorithms. In fact, at the core of link-state algorithms there is a distributed and replicated database. This database is essentially a dynamic map of the whole network, describing the details of all its components and their current interconnections. Using this database as input, each node calculates its best paths using an appropriate algorithm like Dijkstra's (1959) algorithm (a wide variety of alternative efficient algorithms are available, as described for example in Cherkassky, Goldberg, & Radzik, 1994). The memory requirements for each node in this case are $O(N^2)$.

In the most common form of link-state algorithm, each node acts autonomously, broadcasting information about its link costs and states and computing shortest paths from itself to all the destinations on the basis of its local link costs estimates and of the estimates received from other nodes. Each routing information packet is broadcast to all the neighbor nodes that in turn send the packet to their neighbors and so on. A distributed flooding mechanism (Bertsekas & Gallager, 1992) supervises this information transmission trying to minimize the number of re-transmissions.

As in the case of vector-distance, the described algorithm is a general template and a variety of different versions have been implemented to make the algorithm behavior more robust and efficient (Moy, 1998).